\documentclass[pdflatex,sn-mathphys-num,iicol]{sn-jnl}

\usepackage{fontawesome5}
\usepackage{subcaption}
\usepackage{multirow}
\usepackage{array}
\usepackage{graphicx}%
\usepackage{multirow}%
\usepackage{amsmath,amssymb,amsfonts}%
\usepackage{amsthm}%
\usepackage{mathrsfs}%
\usepackage[title]{appendix}%
\usepackage[table]{xcolor}
\usepackage{xcolor}%
\usepackage{textcomp}%
\usepackage{graphicx}
\usepackage{amssymb}
\usepackage{amsmath}
\usepackage{manyfoot}%
\usepackage{booktabs}%
\usepackage{listings}%
\usepackage[nolist]{acronym}
\usepackage{enumitem}
\usepackage[ruled,vlined,linesnumbered]{algorithm2e}
\SetAlgoSkip{} 
\SetAlFnt{\small}
\usepackage[noend]{algpseudocode}
\usepackage{siunitx}

\newtheorem{problem}{Problem}

\newtheorem{remark}{Remark}

\DeclareMathOperator*{\minimize}{\text{minimize}}

\theoremstyle{thmstyleone}%

\theoremstyle{thmstyletwo}%

\theoremstyle{thmstylethree}%

\DeclareMathOperator*{\argmin}{arg\,min}

\title[Perception-Aware Multi-UAV Implicit Coordination in the
Wild]{Perception-Aware Multi-UAV Implicit Coordination in the Wild}

\author*[1,2]{\fnm{Manuel} \sur{Boldrer}}\email{m.boldrer@saxion.nl}
\author[2]{\fnm{Michal} \sur{Kamler}}\email{michal.kamler@fel.cvut.cz}
\author[2]{\fnm{Afzal} \sur{Ahmad}}\email{afzal.ahmad@fel.cvut.cz}
\author[2]{\fnm{Martin} \sur{Saska}}\email{martin.saska@fel.cvut.cz}

\affil[1]{\orgdiv{Department of Life Science, Engineering \& Design}, \orgname{Saxion University of
Applied Sciences}, \orgaddress{\street{Ariënsplein 1 - 300}, \city{Enschede},
  \postcode{7511JX}, \country{The Netherlands}}}
\affil[2]{\orgdiv{Department of Cybernetics}, \orgname{Czech  Technical University in
  Prague}, \orgaddress{\street{ Karlovo namesti 13}, \city{Prague 2},
  \postcode{12135}, \country{Czechia}}}

\begin{acronym}
	\acro{GNSS}[GNSS]{Global Navigation Satellite System}
	\acro{MOCAP}[mo-cap]{Motion capture}
	\acro{MPC}[MPC]{Model Predictive Control}
	\acro{UAV}[UAV]{Unmanned Aerial Vehicle}
	\acro{UV}[UV]{ultraviolet}
	\acro{UVDAR}[\emph{UVDAR}]{UltraViolet Direction And Ranging}
	\acro{RTK}[RTK]{Real-Time Kinematic}
	\acro{ROS}[ROS]{Robot Operating System}
	\acro{LIDAR}[LiDAR]{Light Detection And Ranging}
	\acro{UWB}[UWB]{Ultra-wideband}
	\acro{RBL}[RBL]{Rule-Based Lloyd}
	\acro{IRBL}[iRBL]{improved Rule-Based Lloyd}
	\acro{FOV}[FoV]{Field-of-View}
	\acro{CIRI}[CIRI]{Configuration-space Iterative Regional Inflation}
	\acro{CWVD}[CWVD]{Convex Weighted Voronoi Decomposition}
	\acro{SR}[SR]{Success Rate}
\end{acronym}

\begin{document}

\abstract{
We present a perception-aware 3D navigation system for decentralized multi-robot
coordination in cluttered environments, such as forests with dense canopy cover
where GNSS is unavailable. Our novel perception-aware approach computes
collision-free motion toward a goal region while accounting for the limited
field of view of the sensor, and is combined with a replanner to guide multiple
robots to their goal. The system relies on an onboard anisotropic 3D LiDAR, used
for localization, mapping, and detection of obstacles and neighboring robots.
Decentralized coordination among the robots emerges as a byproduct of local
interactions and common rules, so no explicit communication is required. The system is
evaluated through extensive single and multi-robot simulations across diverse
scenarios and validated in multiple real-world field experiments. 

\begin{small}{\textbf{\textit{Video}---https://mrs.fel.cvut.cz/irbl}}

 \textbf{\textit{Code}---https://github.com/ctu-mrs/irbl}
  \end{small}
} 

\maketitle

\section{Introduction}
\label{sec:Introduction}
In recent years, aerial robots have been successfully deployed in a wide range
of applications, including agriculture, wildlife monitoring, and disaster
response~\cite{papalia2025roadmap}. Across all these domains, the ability to
navigate safely and efficiently through complex environments, while sharing the
space with other robots, is a fundamental requirement. Unstructured
environments with narrow spaces and limited visibility of dynamic obstacles
(other robots), exacerbate the complexity of this navigation problem.


\noindent While single-robot autonomous navigation methods have achieved
impressive high-speed flights in cluttered environments like
forests~\cite{ren2025safety}, the multi-robot setting still remains an open
challenge.
Existing solutions rely on sharing current state (e.g., position and
velocity, like in ORCA-based methods~\cite{van2011reciprocal,kratky2025rvc}) or even future trajectory
information over a communication network~\cite{kondo2026mighty,zhou2022swarm} (e.g. WiFi, UWB).
Common issues like communication delays, dropouts, and unreliable global localization system (GNSS) can lead to collisions between the robots.
As the communication network is a single point of failure it severely limits the scalability of communication-dependent methods to larger
number of robots and reduces the reliability of the overall system.

Removing the reliance on communication can make the method scalable, secure and
fault-tolerant, but brings its own set of challenges.
In~\cite{boldrer2025swarming}, the authors propose an approach that
uses onboard sensors to detect the obstacles and other robots and safely
navigates the multi-robot system to the goal location.
However, the method exhibits three main limitations: (i) the robots are
constrained to planar motion (2D), thus preventing the exploitation of full
three-dimensional mobility; (ii) the robots lack the capability to re-plan their
paths online during the mission, and (iii) the robots motion capabilities are
constrained by the algorithm's conservative speed selection. These limitations
can result in slow flights and potential deadlock conditions in complex
environments that include dense vegetation (e.g., branches, tree crowns or bushes)
or maze-like structures. 



\begin{figure}[t]
   \includegraphics[width=1\columnwidth]{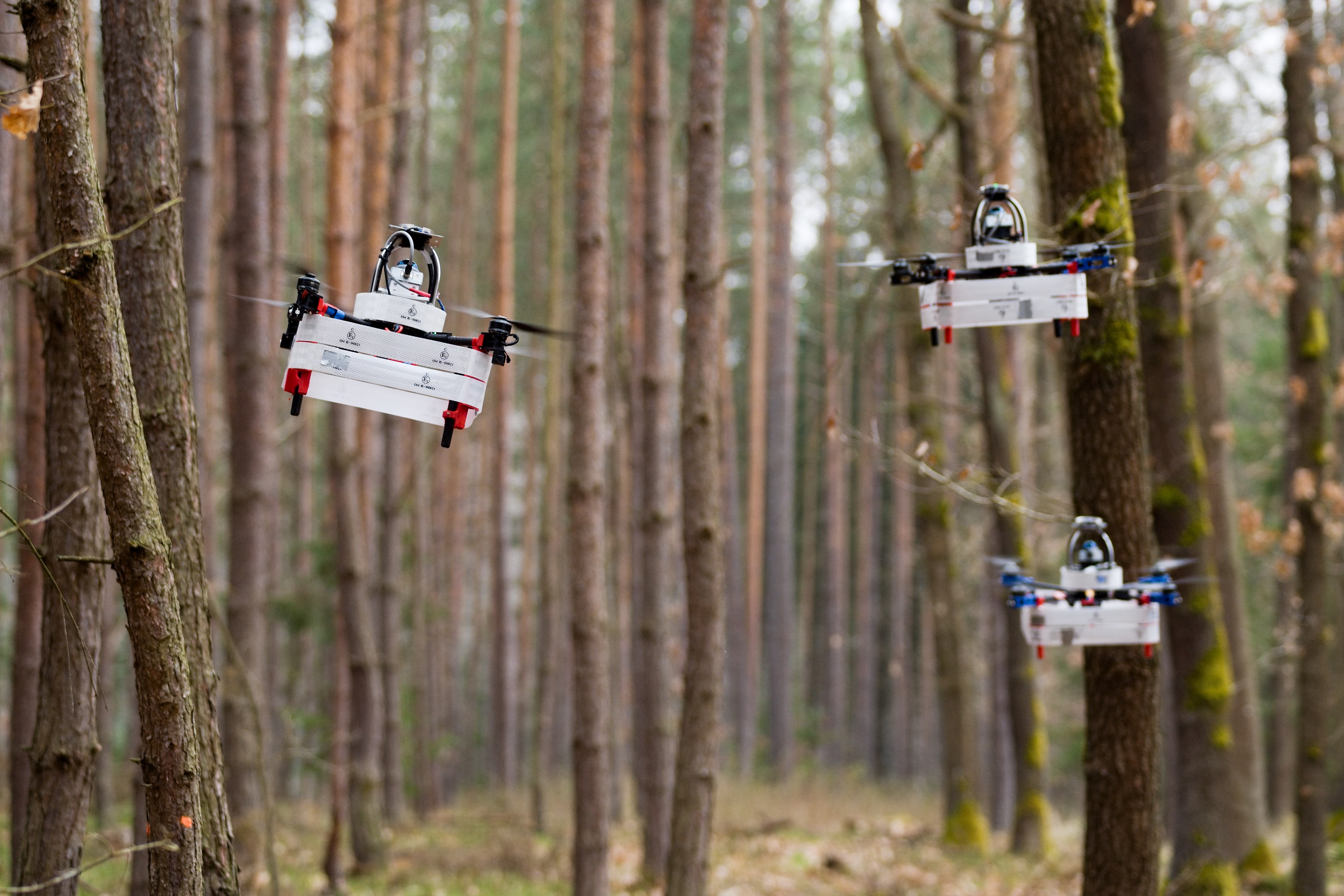} 
   \caption{Snapshot during a real-world experiment in the forest with $3$ UAVs.}
 \label{fig:platform}
\end{figure}
We propose a perception-aware 3D navigation system that accounts for limited field-of-view sensing on each robot and navigates multiple robots in complex environments without any explicit communication.
The robots localize others without exchanging any messages or relying on a shared
communication network. Hence, the robot coordination emerges as a byproduct of local
interactions and common rules, rather than being explicitly enforced.

Our contribution is twofold. i. We propose iRBL, which improves and extends the
Rule-Based Lloyd (RBL) algorithm~\cite{boldrer2025swarming} to fully
three-dimensional environments while incorporating sensing constraints. In
particular, iRBL explicitly accounts for anisotropic and limited field-of-view
perception from onboard 3D LiDAR, enabling perception-aware motion generation
and heading adaptation. The framework further integrates a CIRI-based convex
decomposition~\cite{ren2025safety} for safe-region construction in cluttered 3D
environments, together with an online replanning module that improves
performance and success rate in low-traversability scenarios. ii. A fully
integrated onboard system with extensive real-world validation and fully
open-source software release. We design and implement a complete
communication-free navigation pipeline that integrates iRBL with MPC-based
trajectory tracking, Point-LIO state estimation, onboard mapping, and
LiDAR-based inter-robot detection. The resulting system operates fully onboard
in GNSS-denied forest environments without external infrastructure. Extensive
simulations and real-world experiments demonstrate that the proposed system
achieves a higher success rate than state-of-the-art
methods~\cite{boldrer2025swarming,zhou2022swarm}: in the most cluttered
environments ($\tau \approx 1$~m), \ac{IRBL} reaches the goal in all runs,
whereas RBL~\cite{boldrer2025swarming} and EGO2~\cite{zhou2022swarm} succeed in
at most $20\%$ of the runs. In multi-robot scenarios, \ac{IRBL} achieves up to
$2.7$ times the average speed of RBL, and it remains successful in cluttered
environments where RBL fails. To support reproducibility and further research,
we provide the software implementation of the proposed framework.

\subsection{Related Work}
\label{sec:RelatedWork}
\begin{table*}[t]

\renewcommand{\tabcolsep}{0.06cm}
\centering
\footnotesize

\rowcolors{2}{gray!8}{white}

\begin{tabular}{c c c c c c c c c c c c}
\toprule

& EGO2\cite{zhou2022swarm} &
RMADER\cite{kondo2023robust} &
PACNAV\cite{ahmad2022pacnav} &
RBL\cite{boldrer2025swarming} &
HDSM\cite{toumieh2024high} &
RVC-NMPC\cite{kratky2025rvc} &
Chen\cite{chen2023multi} &
DREAM\cite{csenbacslar2024dream} &
AMSWARM\cite{adajania2023amswarm} &
Choi\cite{choi2026communication} &
iRBL \\

\midrule

\textcolor{black!60}{\faCommentSlash} &
\textcolor{black!60}{\faTimes} &
\textcolor{black!60}{\faTimes} &
\textcolor{black!60}{\faCheck} &
\textcolor{black!60}{\faCheck} &
\textcolor{black!60}{\faTimes} &
\textcolor{black!60}{\faTimes} &
\textcolor{black!60}{\faTimes} &
\textcolor{black!60}{\faTimes} &
\textcolor{black!60}{\faTimes} &
\textcolor{black!60}{\faCheck} &
\textcolor{black!60}{\faCheck} \\

\textcolor{black!60}{\faEye} &
\textcolor{black!60}{\faDotCircle} &
\textcolor{black!60}{\faTimes} &
\textcolor{black!60}{\faDotCircle} &
\textcolor{black!60}{\faDotCircle} &
\textcolor{black!60}{\faTimes} &
\textcolor{black!60}{\faTimes} &
\textcolor{black!60}{\faTimes} &
\textcolor{black!60}{\faTimes} &
\textcolor{black!60}{\faTimes} &
\textcolor{black!60}{\faDotCircle} &
\textcolor{black!60}{\faCheck} \\

\textcolor{black!60}{\faTree} &
\textcolor{black!60}{\faCheck} &
\textcolor{black!60}{\faTimes} &
\textcolor{black!60}{\faCheck} &
\textcolor{black!60}{\faCheck} &
\textcolor{black!60}{\faTimes} &
\textcolor{black!60}{\faDotCircle} &
\textcolor{black!60}{\faTimes} &
\textcolor{black!60}{\faTimes} &
\textcolor{black!60}{\faTimes} &
\textcolor{black!60}{\faCheck} &
\textcolor{black!60}{\faCheck} \\

\textcolor{black!60}{\faStopwatch} &
\textcolor{black!60}{\faArrowUp} &
\textcolor{black!60}{\faArrowUp} &
\textcolor{black!60}{\faAngleDoubleDown} &
\textcolor{black!60}{\faArrowDown} &
\textcolor{black!60}{\faArrowUp} &
\textcolor{black!60}{\faAngleDoubleUp} &
\textcolor{black!60}{\faArrowUp} &
\textcolor{black!60}{\faArrowUp} &
\textcolor{black!60}{\faArrowUp} &
\textcolor{black!60}{\faArrowDown} &
\textcolor{black!60}{\faArrowUp} \\

\textcolor{black!60}{\faGlobe} &
\textcolor{black!60}{\faCheck} &
\textcolor{black!60}{\faCheck} &
\textcolor{black!60}{\faCheck} &
\textcolor{black!60}{\faCheck} &
\textcolor{black!60}{\faCheck} &
\textcolor{black!60}{\faTimes} &
\textcolor{black!60}{\faCheck} &
\textcolor{black!60}{\faTimes} &
\textcolor{black!60}{\faCheck} &
\textcolor{black!60}{\faTimes} &
\textcolor{black!60}{\faCheck} \\

\bottomrule
\end{tabular}

\vspace{6pt}
\caption{Qualitative comparison with state-of-the-art algorithms on the aspects
of interest (i.e., implicit coordination
(\textcolor{black!60}{\faCommentSlash}), anisotropic perception
(\textcolor{black!60}{\faEye}), test in the wild
(\textcolor{black!60}{\faTree}), operational speed
(\textcolor{black!60}{\faStopwatch}) and open-source code
(\textcolor{black!60}{\faGlobe})). We indicate with
``\textcolor{black!60}{\faTimes}'' if the algorithm does not implement the
feature, ``\textcolor{black!60}{\faCheck}'' if it does, and
``\textcolor{black!60}{\faDotCircle}'' if it does only partially. In the fourth
row the operational speed is indicated with arrows, where
``\textcolor{black!60}{\faAngleDoubleDown}'' indicates poor performance,
``\textcolor{black!60}{\faAngleDoubleUp}'' indicates high performance, while
``\textcolor{black!60}{\faArrowDown}'' and ``\textcolor{black!60}{\faArrowUp}''
indicate intermediate low and intermediate high performance, respectively.}
\label{tab:sota}
\end{table*}
Modern methods enable a single robot to operate safely and efficiently in
relatively complex
environments~\cite{ren2022bubble,zhou2021raptor,ren2025safety,pochobradsky2025geometric}.
However, solutions for coordinating multiple robots in such environments remain
limited. Popular approaches rely on a communication
network~\cite{zhou2022swarm,kondo2026mighty,kondo2023robust,yin2025visibility,kratky2025rvc} to
share unobservable information and use a common reference frame among the robots
for safe motion planning. These approaches, however, are not always practical in
the real-world scenarios and often require a level of reliability of the
communication network that is difficult to achieve. Motivated by these
challenges, communication-free reactive methods have also been
proposed~\cite{boldrer2025swarming,ahmad2022pacnav,choi2026communication} and
validated in challenging environments. However, these approaches still exhibit
limitations, particularly in crowded scenarios, where performance and mission
success rate can degrade significantly. Our approach eliminates the need for
inter-robot communication while advancing multi-robot navigation capabilities in
complex environments. Specifically, we show that the improved rule-based Lloyd
algorithm (iRBL) provides a safe and reliable solution that achieves a higher success rate than
optimization-based methods such as~\cite{zhou2022swarm} in challenging settings.
We see our contributions as a key step towards robust, implicit multi-robot
coordination in the wild, where the coordination naturally emerges from the robot
behavior rather than through explicit message exchange. In Table~\ref{tab:sota},
we provide a qualitative comparison with the most relevant approaches across the
following aspects:
\begin{itemize}
  \item \textit{Implicit Coordination} (\textcolor{black!60}{\faCommentSlash}):
    the approach supports decentralized coordination without explicit
    inter-agent communication;
  \item \textit{Anisotropic Perception} (\textcolor{black!60}{\faEye}): the system relies on
      perception to avoid obstacles and other robots, and it explicitly accounts for
      limited field-of-view;
    \item \textit{Testing in the Wild} (\textcolor{black!60}{\faTree}): the algorithm has been
      evaluated in real-world and GNSS-denied environments, as opposed to simulation-only or
      motion-capture based experiments;
    \item \textit{Operational Speed} (\textcolor{black!60}{\faStopwatch}): The average speed maintained during the mission.
      \item \textit{Open Source} (\textcolor{black!60}{\faGlobe}): The
        algorithm is fully open-source.
\end{itemize}
This representation enables quick visual comparison across approaches,
emphasizing qualitative aspects important for real-world UAV deployments.






\section{Problem Description}
\label{sec:problem description}

This article addresses the following problem. 
\begin{problem}
  Given a group of $N$ aerial robots equipped with limited \ac{FOV} sensors, safely navigate them to a
  goal region in an unknown environment. The robots can only use onboard sensors and computational capabilities and cannot 
  share any information with other robots.
\label{pr:main}
\end{problem}

\noindent In the presented problem, each aerial robot uses an anisotropic sensor, such as a 3D \ac{LIDAR}, to localize itself and detect obstacles and other agents in its surroundings.
Due to the restricted \ac{FOV} of the \ac{LIDAR}, the aerial robots have a limited knowledge of the collision-free space around them.
This free space is further reduced when other robots are also operating in the same region.
The limited \ac{FOV} along with other moving robots severely constrains the collision-free space around the aerial robots and limits their motion when flying in cluttered, unknown environments.


\section{Perception-Aware Motion Planning}
\label{sec:approach} 

We propose a perception-aware 3D motion planning approach, \ac{IRBL}. The \ac{IRBL} builds upon \ac{RBL}~\cite{boldrer2025swarming} and relies solely on onboard sensing and computational capabilities for collision-free motion planning, thereby forming a decentralized system in which coordination among the robots emerges without explicit communication.
However, unlike \ac{RBL}~\cite{boldrer2025swarming}, the proposed \ac{IRBL} incorporates active perception to explicitly account for the limited \ac{FOV} to improve the safety and speed when flying in environments with low traversability.
Furthermore, \ac{IRBL} employs a \acf{CIRI}-based~\cite{ren2025safety} convex decomposition to find a collision-free space directly using the pointcloud and extends \ac{CWVD} to account for surrounding aerial robots when moving in 3D.
The \ac{IRBL} algorithm executed by each robot $i$ is summarized as follows:
\begin{itemize}
  \item[i.] Compute a safe visible region $\mathcal{W}_i \subseteq
    \mathbb{R}^3$ using a safe region $\mathcal{B}_i \subseteq
    \mathbb{R}^3$ and the sensor \ac{FOV} i.e.,
    $\mathcal{W}_i = \mathcal{B}_i \cap FOV_i$, such that the robot position $p_i
    \in \mathcal{W}_i$;
  \item[ii.] Using a desired density function $\psi_i(q)$ : $\mathbb{R}^3 \rightarrow \mathbb{R}^{\geq 0}$, compute the centroid position $$c_{\mathcal{B}_i} = \frac{\sum_{q
    \in \mathcal{B}_i} q \psi_i(q)}{\sum_{q \in \mathcal{B}_i}
    \psi_i(q)}, $$
    where, in practice, $\mathcal{B}_i$ is discretized into a finite set of
    points over which the sums are computed;
  \item[iii.] Compute the projection of the centroid position $c_{\mathcal{B}_i}$
    onto the set $\mathcal{W}_i$
    $$\operatorname{proj}_{\mathcal{W}_i}(c_{\mathcal{B}_i}) = \arg\min_{y \in
    \mathcal{W}_i} \|c_{\mathcal{B}_i} - y\|
$$
\item[iv.] Steer the robot $i$ to the position
  $\operatorname{proj}_{\mathcal{W}_i}(c_{\mathcal{B}_i})$ if $\langle 
    h_i^D, h_i \rangle > \cos(f_x/2)$, where $h_i$ is the heading unit vector
    of the robot in the $xy$-plane, $h_i^D$ is the unit vector pointing from
    $p_i$ towards $c_{\mathcal{B}_i}$ in the $xy$-plane, and $f_x$ is the
    horizontal field-of-view (see Figure~\ref{fig:fov}), otherwise adjust the
    heading towards $c_{\mathcal{B}_i}$.

\end{itemize}
The above procedure defines the core steps of the proposed iRBL algorithm, which
is executed independently by each robot only using locally observable
information. The behavior of each robot is determined by the sets
$\mathcal{B}_i$ and $\mathcal{W}_i$ and the density function $\psi_i(q)$, which
jointly specify a desired point to be tracked at high frequency by the
underlying low-level controller.


\subsection{Computing the set \texorpdfstring{$\mathcal{B}_i$}{B\_i} and \texorpdfstring{$\mathcal{W}_i$}{W\_i}}

To determine the centroid position $c_{\mathcal{B}_i}$, we first 
define the set $\mathcal{B}_i = \{\mathcal{S}_i \cap \mathcal{A}_i \cap
\mathcal{C}_i\}$. The set $\mathcal{S}_i = \{q \in \mathbb{R}^3 \mid \|q-p_i\|<
r_{s,i}\}$ is the set of all points within a distance $r_{s,i}$ from the current
position of the robot $p_i$. 
Set $\mathcal{S}_i$ restricts the centroid to lie within the distance $r_{s,i}$ of the robot thereby controlling the planning horizon.
The set $\mathcal{S}_i$ does not account for other robots or 
obstacles thus specifying the largest set of nearby points that can be selected to
compute the centroid.

In order to prevent collisions with other robots, we compute the set of collision-free points $\mathcal{A}_i$ that accounts for the nearby robots.
We extend the \acf{CWVD}~\cite{boldrer2025swarming}
method to 3D space to generate the set $\mathcal{A}_i$ as follows.
We first define two half-spaces $\mathcal{H}_{ij}^1$ and
$\mathcal{H}_{ij}^2$, such that
\begin{equation}
  \label{eq:halfspaces}
  \begin{split}
    \mathcal{H}_{ij}^{1} &=
  \left\{ q \in \mathbb{R}^3 \mid
    n_{p_{ij}} \cdot (q - a_{p_{ij}})
  \leq 0 \right\}, \\
  \mathcal{H}_{ij}^{2} &=
  \left\{ q \in \mathbb{R}^3 \mid
    n_{p_{ij}} \cdot (q - \tilde{p}_j)
  > 0 \right\},
  \end{split}
\end{equation}
where $n_{{p}_{ij}} = \tilde{p}_{j} - \tilde{p}_{i}$ defines the normal to the
hyperplane separating the robots $i$ and $j$, while $a_{p_{ij}} =
\tilde{p}_{i}+\epsilon \ n_{p_{ij}}$ is a point along the normal, with $\epsilon \in [0, 1]$.
The points $\tilde{p}_{i}$ and $\tilde{p}_{j}$  are given by
\begin{align} \tilde{p}_{i} &= p_i +\Delta_{i,j}\frac{p_{j} - p_{i}}{\| p_{j} -
p_{i} \|} , & \tilde{p}_{j} &= p_{j}+ \Delta_{i,j}\frac{p_{i} - p_{j}}{\| p_{i}
- p_{j} \|}, \end{align} where $\delta_{i}$ and $\delta_{j}$ are the radii of respective robots and $\Delta_{i,j}= \delta_{i} + \delta_{j}$.
The set associated with the $i$-th robot with respect to the $j$-th robot is then computed as 
\begin{equation}
  \label{eq:half}
  \mathcal{A}_{ij} =
  \begin{cases}
    \mathcal{H}_{ij}^{1} & \text{if }
  \| \tilde{p}_{i} - p_i \|
  \leq \| \tilde{p}_{j} - p_i \|, \\
    \mathcal{H}_{ij}^{2} & \text{otherwise}.
  \end{cases}
\end{equation}

\begin{figure}[t]
 \centering
 \setlength{\tabcolsep}{0.05em}
 \begin{tabular}{cc}
   \includegraphics[width=0.5\columnwidth]{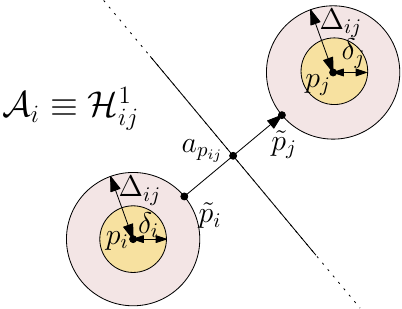} &
   \includegraphics[width=0.5\columnwidth]{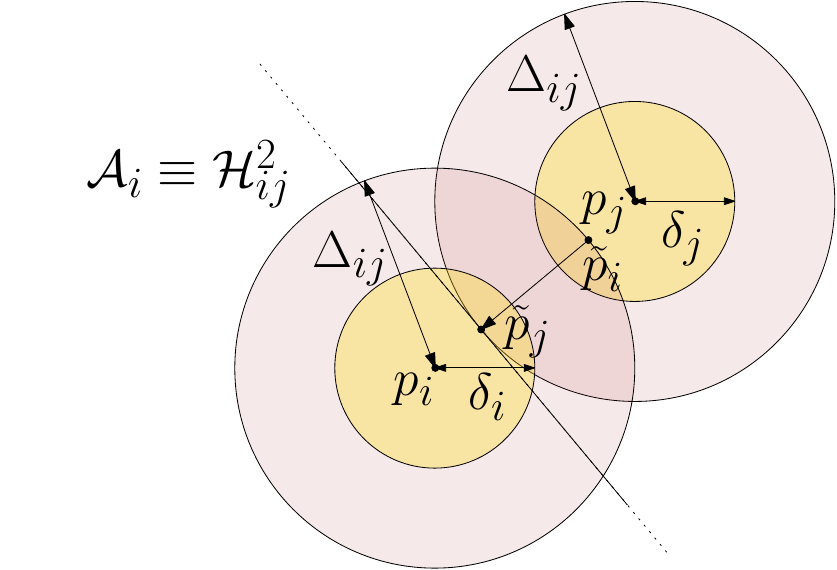} 
   \end{tabular}
   \caption{2D Graphical representation of the set $\mathcal{A}_i$. On the left,
   the case where the robots are separated by more than $2\Delta_{ij}$; on the
   right, the case where they are separated by less than $2\Delta_{ij}$. The
   half-spaces are constructed according to~\eqref{eq:halfspaces} such that all
   the points are collision-free.}
  \label{fig:sets}
\end{figure}

\noindent Finally, the set $\mathcal{A}_i$ accounts for all the neighboring
robots, i.e., $\mathcal{A}_i = \bigcap_{j \in \mathcal{N}_i} \mathcal{A}_{ij}$,
where $\mathcal{N}_i$ denotes the set of robots detected by the $i$-th robot.
The intersection of sets $\mathcal{S}_i$ and $\mathcal{A}_i$ contains
all the points within distance $r_{s,i}$ that are safe from collisions into surrounding robots.
Figure~\ref{fig:sets} provides a two-dimensional illustration of the set $\mathcal{A}_{ij}$~\eqref{eq:half}. 

\begin{figure}[t]
 \centering
   \includegraphics[width=1\columnwidth]{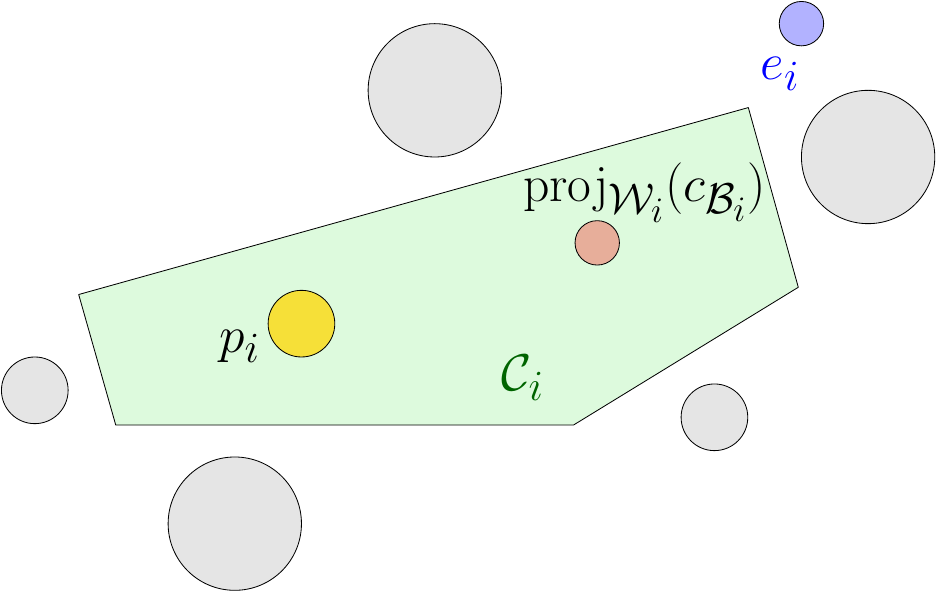} 
   \caption{2D Graphical representation of the set $\mathcal{C}_i$ (which in
   this particular case correspond also with the set $\mathcal{B}_i$) computed
   using the proposed seeds. The obstacles are represented by grey circles.}
   \label{fig:setC}
\end{figure}

The extended \ac{CWVD} method is designed to obtain collision-free half-spaces for moving robots but can not be directly used to find collision-free space when a robot is surrounded by several obstacles.
To avoid collisions with static obstacles, we use the \ac{CIRI} method~\cite{ren2025safety} which computes the collision-free space directly using the pointcloud around a robot.
\ac{CIRI} operates on the obstacle pointcloud (other robots
are removed from the pointcloud), and represents each input point as a sphere
with radius matching the radius of the $i$-th robot. It selects two points
$\mathcal{D}_i=\{s_a, s_b\}$, which are used for the generation of the convex
decomposition. Starting from the pointcloud and the seeds, CIRI computes the
convex decomposition of the configuration space by iteratively inflating an
ellipsoid, parameterized by the focal set $\mathcal{D}_i$, until it reaches
collision with an obstacle. By directly operating on the obstacle pointcloud, it
provides a computationally efficient way of computing a collision-free convex
decomposition $\mathcal{C}_i$ of the space around the $i$-th robot.
At the first iteration, we initialize CIRI with the robot position as
the seeds, $\mathcal{D}_i=\{p_i,p_i\}$. At each subsequent iteration, we use
$\mathcal{D}_i=\{p_i,\text{proj}_{\mathcal{W}_i}(c_{\mathcal{B}_i})\}$, where
$\text{proj}_{\mathcal{W}_i}(c_{\mathcal{B}_i})$ is the value computed at the
previous iteration.
Using $\text{proj}_{\mathcal{W}_i}(c_{\mathcal{B}_i})$ instead of just the centroid $c_{\mathcal{B}_i}$, biases the convex decomposition towards the previously computed $\text{proj}_{\mathcal{W}_i}(c_{\mathcal{B}_i})$ thus generating more collision-free points around it.
In comparison to the CWVD method, CIRI can generate a convex
decomposition with a desired directionality, which is more suitable and
effective for navigation towards a goal. In Figure~\ref{fig:setC}, we provide a
two-dimensional graphical representation of the set $\mathcal{C}_i$ computed using \ac{CIRI}.

\begin{figure}[t]
 \centering
   \includegraphics[width=1\columnwidth]{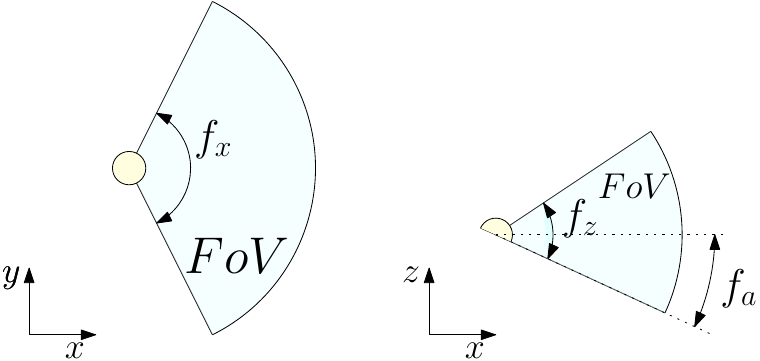} 
   \caption{Graphical representation of the sensor field-of-view $\mathrm{FoV} =
   \{f_x, f_z, f_a\}~(\deg)$. }
   \label{fig:fov}
\end{figure}

The safe visible set
$\mathcal{W}_i$ is defined as the intersection between the set $\mathcal{B}_i= \{\mathcal{S}_i \cap \mathcal{A}_i \cap
\mathcal{C}_i\}$
and the set of points within the robot \ac{FOV} i.e., $\mathcal{W}_i
= \mathcal{B}_i \cap \mathrm{FoV}_i$. We define the field-of-view as $
\mathrm{FoV} = \{f_x, f_z, f_a\}~(\deg)$, where \(f_x\) and \(f_z\) represent
the horizontal and vertical FoV, respectively, and \(f_a\) denotes the sensor
pitch angle relative to the robot body frame (see Fig.~\ref{fig:fov}). To
ensure that $\mathcal{W}_i$ is convex, we restrict the horizontal field-of-view
to $f_x \le 180^\circ$. Hence, even when the sensor provides $360^\circ$
horizontal coverage, as the Livox MID-360 used in our experiments does, only its
forward half is used.

\begin{remark}
  The need for different set definitions for static and dynamic obstacles arises
  from safety considerations. In principle, the CIRI formulation can be applied
  to both static and dynamic obstacles; however, doing so would lead to overly
  aggressive (and selfish) robot behavior. Hence, we choose to combine CWVD for
  dynamic obstacles (other robots) and CIRI for static obstacles, to enhance
  multi-robot coordination. A quantitative comparison is provided in the
  simulation results section.
\end{remark}

\subsection{The density function \texorpdfstring{$\psi_i(q)$}{psi\_i(q)}}
The density function $\psi_i(q)$ assigns a weight to each point $q$ in the safe region
$\mathcal{B}_i$ and is used to compute the centroid $c_{\mathcal{B}_i}$.
As shown in previous studies~\cite{boldrer2023rule, boldrer2025swarming}, the
design of the density function plays a crucial role in decreasing the time to reach the
goal region and dictates the implicit coordination between the robots.
In the proposed density function $\psi_i(q)$, the points $q \in \mathcal{B}_i$ are weighted according to their distance from the 
point of interest $\bar{p}_i$. By assigning higher weights to points closer to
the point of interest, the density function biases the centroid toward the $\bar{p}_i$, thereby
driving the robot in that direction. Inspired by~\cite{boldrer2023rule}, we define $\psi_i(q)$ as

\begin{equation}\label{eq:psi}
  \psi_i(q) =
  \text{exp}{ \left( -\frac{\|q-\bar{p}_i\|}{\beta_i}\right)}.
  \end{equation}
  The time-varying spreading factor $\beta_i \in \mathbb{R}$ evolves according to \begin{equation}
    \label{eq:rho}
    \dot{\beta}_i = \begin{cases} -k_{\beta}\beta_i \,\,\,
      &\text{if}\,\, \|c_{\mathcal{B}_i}-p_i\| <d_{1} \, \land \,\\
      &\|c_{\mathcal{B}_i} -c_{\mathcal{S}_i}\|> d_{2}, \\
      -k_{\beta}(\beta_i-\beta^D_i) &\text{otherwise,} \end{cases}
  \end{equation}
  where, $c_{\mathcal{S}_i} \in \mathcal{S}_i$ is the centroid of set
  $\mathcal{S}_i$ and $\beta_i^D, d_1, d_2, k_{\beta}$ are positive constants.
  As the robots often suffer from tracking errors due to model mismatch and
  external disturbances, and the estimation of the surrounding robot’s pose may
  not be precise, we limit $\beta_i$ using
  \begin{equation}\label{eq:beta_min} \beta^{\min}_i = \argmin_{\beta_i}
    \left(\|c_{\mathcal{B}_i} - p_{\partial \mathcal{B}_i} \| - d_u\right)^2,
  \end{equation} where $\partial \mathcal{B}_i$ represents the contour of the
  set $\mathcal{B}_i$, $p_{\partial \mathcal{B}_i}$ is the point of
  $\partial \mathcal{B}_i$ closest to $c_{\mathcal{B}_i}$, and $\beta^{\min}_i$
  is the minimum admissible value of $\beta_i$, i.e., $\beta_i \geq
  \beta^{\min}_i$ at all times. Here, $d_u \geq 0$ is the parameter that quantifies the
  level of uncertainties in the system, i.e., the sum of the measurement
  uncertainties and the tracking error (refer to~\cite{boldrer2025swarming} for
  more details).

  The local point of interest $\bar{p}_i \in \mathbb{R}^3$ guides the robot towards the goal region and evolves according to
\begin{equation}
  \label{eq:wp}
  \begin{aligned}
    \dot{\bar{p}}_i &=
    \begin{cases}
      -k_{\mathrm{wp}}\bigl(\bar{p}_i - \mathrm{wp}_i^{\varepsilon}\bigr), & \text{if } \|c_{\mathcal{B}_i}-p_i\| < d_3\\
        & \land\ \|c_{\mathcal{B}_i}-c_{\mathcal{S}_i}\| > d_4,\\
      -k_{\mathrm{wp}}\bigl(\bar{p}_i - \mathrm{wp}_i\bigr), & \text{otherwise,}
    \end{cases}\\[3pt]
    \bar{p}_i &\leftarrow \mathrm{wp}_i \quad \text{if }
    \begin{aligned}[t]
      &\|p_i-\bar{c}_{\mathcal{B}_i}\| > \|p_i-c_{\mathcal{B}_i}\|\\
      &\land\ \bar{p}_i = \mathrm{wp}_i^{\varepsilon},
    \end{aligned}
  \end{aligned}
\end{equation}
where, $d_3, d_4$, and $k_{\mathrm{wp}}$ are positive constants. The matrix $R_\varepsilon=R(\tfrac{\pi}{2}-\varepsilon)$
denotes an $xy$-planar rotation, $\mathrm{wp}_i$ represents the desired waypoint
  location of the $i$-th robot, and $\mathrm{wp}_i^{\varepsilon} = p_i +
  R_\varepsilon(\mathrm{wp}_i - p_i)$ is the waypoint rotated about the robot
  position $p_i$. $\varepsilon$ is a small offset introduced to prevent rotations exactly equal
  to $\pi/2$. $\bar{c}_{\mathcal{B}_i}$ corresponds to the
  centroid computed over the cell $\mathcal{B}_i$ when $\bar{p}_i \equiv wp_i$.

The update laws in~\eqref{eq:rho},~\eqref{eq:beta_min}, and~\eqref{eq:wp} are adaptive
rules controlling the behavior of individual robots. The rules in~\eqref{eq:rho} and~\eqref{eq:beta_min} modulate how aggressively robot $i$ advances
toward its target, whereas~\eqref{eq:wp} enforces a consistent right- or
left-hand motion preference among all robots.
When applied to a group of robots, these rules enable decentralized coordination by controlling the robots' direction of motion and speed under conditions of strong symmetric congestion.

\section{System Architecture}
\label{sec:approach_2}
We propose a system for 3D navigation of a group of aerial robots in a highly cluttered environment without exchanging any information among them.
Our system only uses the on-board 3D \ac{LIDAR} for localization~\cite{he2023point}, mapping, and detecting other robots in the surrounding region.
The point cloud from the \ac{LIDAR} is directly used by the proposed \ac{IRBL} algorithm to compute the centroid, which guides each robot's motion.
Using the proposed system on a group of robots yields a decentralized multi-robot system in which coordination emerges without explicit communication. 

\begin{figure}[t]
 \centering
 \setlength{\tabcolsep}{0.05em}
  \includegraphics[width=1\columnwidth]{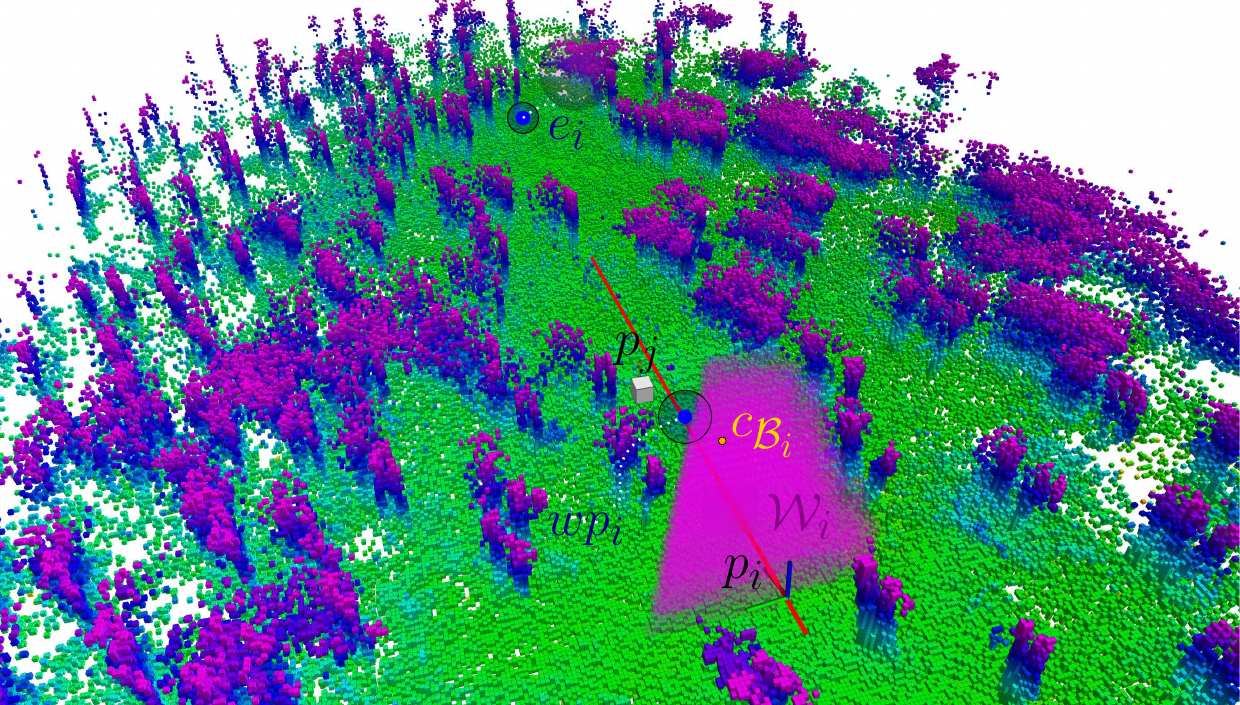}
  \caption{Illustration of the main variables associated with the $i$-th robot
  during a real-world experiment. The robot’s pose $p_i$ is shown as an axis
  reference frame. The set $\mathcal{W}_i$ is represented by the magenta set,
  and the centroid position $c_{\mathcal{B}_i}$ is shown in orange. The planned
  path is depicted by a red line. The $wp_i$ and $e_i$ indicate respectively the
  active waypoint and the final goal positions. Other robots positions $p_j$
  are indicated by a grey box. The colored point cloud illustrates obstacles,
  with different colors corresponding to different z-values.}
  \label{fig:rviz}
\end{figure}
As discussed in Section~\ref{sec:approach}, once we generate the sets $\mathcal{B}_i$ and $\mathcal{W}_i$, we can compute the centroid position $c_{\mathcal{B}_i}$ for the robot $i$.
However, the direction and speed of the robot are also determined by the point of interest $\bar{p}_i$, which evolves based on the desired waypoints obtained from a high-level path planner.
In contrast to~\cite{boldrer2022multi} and~\cite{boldrer2025swarming}, we determine $\bar{p}_i$ from the time-varying path produced by an $A^*$ algorithm~\cite{hart1968formal}.
Although the $A^*$ path planner produces a consistent collision-free path to a goal point $e_i$ through the map generated by Point-LIO~\cite{he2023point}, it does not account for surrounding moving robots.
The proposed system uses the $ A^*$-generated waypoints only as a guiding path. It employs the proposed \ac{IRBL} method to safely navigate, resolve deadlocks, and avoid oscillations.
Figure~\ref{fig:rviz} depicts a real-world experiment illustrating the perception-aware motion planning using the proposed system from the perspective of the $i$-th robot.

\begin{figure}
    \centering
     \includegraphics[width=1\columnwidth]{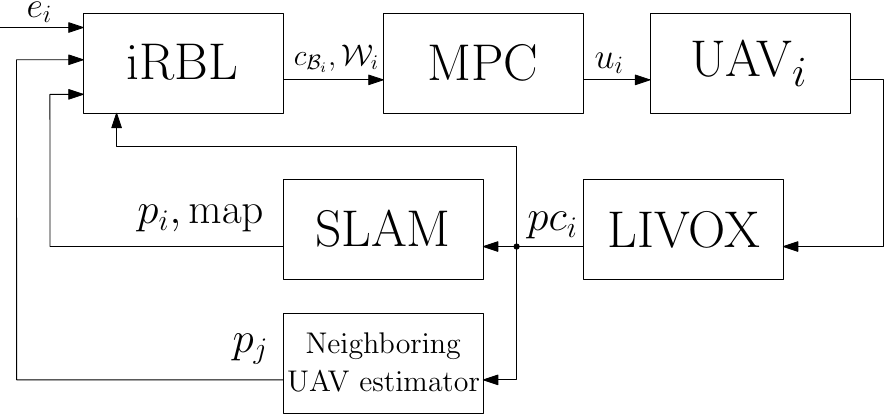} 
     \caption{Block diagram representation of the proposed pipeline. }
   \label{fig:block}
  \end{figure}

Each robot $i$ independently solves a model predictive control problem~\cite{baca2021mrs} to generate control inputs from the obtained centroid position. 
For notational convenience, the robot index is omitted in the following.
The optimization problem is defined as
  \begin{subequations}\label{eq:MPC_problem_rephrased}
    \begin{align}
      \minimize_{\{x_k,u_k\}} \quad & \sum_{k=0}^{N_t}
      J(x_k,h_k,u_k,c_{\mathcal{B}},\mathcal{W}), \label{eq:cost_rephrased}\\
      \textrm{s.t.: } \nonumber \\
      & x_k = f(x_{k-1},u_{k-1}), \quad k = 1,\dots,N_t, \label{eq:dyn_rephrased}\\
      & x_k \in \mathcal{X}, \; u_k \in \mathcal{U}, \quad k = 0,\dots,N_t, \label{eq:bounds_rephrased}\\
      & x_0 = x_{\textrm{init}}, \label{eq:init_rephrased}
    \end{align}
  \end{subequations}
  where $k$ denotes the discrete-time index and $N_t$ is the prediction horizon.
  The state is defined as $x_k =
  [p_k^\top,\dot{p}_k^\top,\ddot{p}_k^\top,\phi_k]^\top$, where $\phi_k$ is the
  yaw angle and $h_k = [\cos\phi_k, \sin\phi_k]^\top$ is the corresponding heading
  unit vector in the $xy$-plane, and
  $u_k$ represents the control input. The function $f(\cdot)$ describes the
  robot dynamics, $\mathcal{X}$ and $\mathcal{U}$ denote the admissible state
  and input sets, respectively, and $x_0 = x_{\textrm{init}}$ specifies the
  initial condition. The cost function $J(\cdot)$ is designed to reduce the
  error with respect to the desired position $e_{p,k} = p_k-p_k^D$ and heading
  $e_{h,k} = h_k-h_k^D$, while
  simultaneously regularizing the control effort.
  Using the centroid obtained from the proposed \ac{IRBL}, we set the
  desired heading as  
  \begin{equation}\label{eq:desired_heading}
    h_k^D= \frac{P(c_{\mathcal{B}}- p_0)}{\|P(c_{\mathcal{B}}-p_0)\|},
  \end{equation}
    where $P = [I_2 \;\; 0_{2\times 1}]$ projects a vector onto the $xy$-plane,
    and $p_0$ is the current robot position. The desired position is set as
  \begin{equation}\label{eq:desired_position}
    p_{k}^D = \begin{cases} \operatorname{proj}_{\mathcal{W}}(c_{\mathcal{B}})
    \,\,\, &\text{if} \,\,\, \langle h_0,
  h_k^D \rangle > \cos(f_x/2),\\
    p_0 \,\,\, &\text{otherwise,} 
  \end{cases}
  \end{equation}
  where $h_0$ is the current heading of the robot.
  The heading adjustment in~\eqref{eq:desired_heading} ensures that the robot actively looks towards the centroid, thereby improving perception of the surroundings while remaining biased towards the goal region.
  The desired position in~\eqref{eq:desired_position} is also selected based on the \ac{FOV}, which further improves visibility of the environment and ensures collision-free motion even when surrounded by other moving robots.
  In Figure~\ref{fig:block}, we provide a graphical representation of the
  overall block diagram for the proposed system.

\begin{remark}\label{re:safety}
  By considering the set $\mathcal{B}_i = \mathcal{S}_i \cap \mathcal{A}_i \cap
  \mathcal{C}_i$, the density function $\psi_i(q)$ defined in~\eqref{eq:psi},
  and computing the control inputs according to the MPC formulation
  in~\eqref{eq:MPC_problem_rephrased}, the i-th robot can navigate safely,
  provided that $d_u$ is chosen sufficiently large to account for uncertainties
  and tracking errors.

  This behavior follows from the following observations. The set $\mathcal{W}_i$
  is convex and contains both the robot position $p_i$ and
  $\operatorname{proj}_{\mathcal{W}_i}(c_{\mathcal{B}_i})$. Consequently, it
  also contains the entire line segment connecting these two points. Under the
  assumptions of perfect reference tracking and no sensing errors, this property
  is sufficient to guarantee that the $i$-th robot remains within the safe set
  during its motion.

  However, in practice, uncertainties such as tracking errors and mutual
  localization errors are always in place. Therefore,
  $d_u$ in~\eqref{eq:beta_min} must be appropriately tuned to provide sufficient
  margin for these uncertainties and preserve safe behavior. A formal safety
  proof is beyond the scope of this work; nevertheless, the following intuitive
  argument provides insight into the role of the parameter $d_u$.
  \begin{figure}[t]
   \centering
     \includegraphics[width=1\columnwidth]{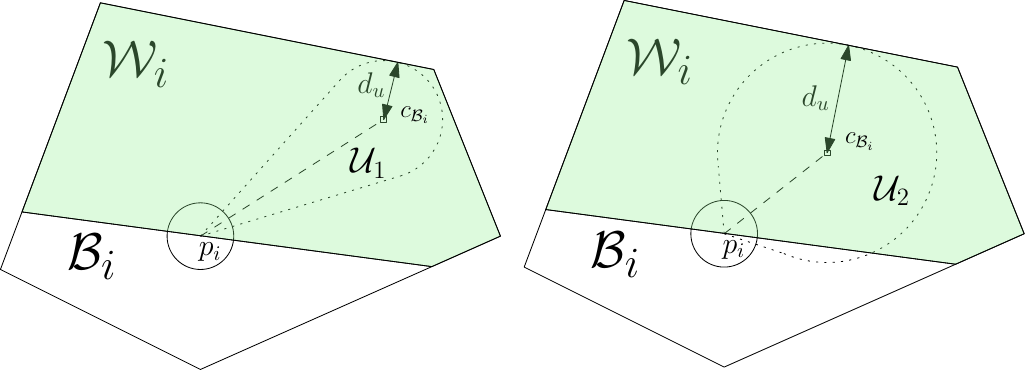} 
     \caption{Top view illustration of the effect of the parameter $d_u$.}
     \label{fig:safety}
  \end{figure}
  In Figure~\ref{fig:safety} we show the effect of a smaller (left) and a
  larger (right) value of $d_u$. It can be noticed, through geometrical
  construction, that choosing higher values of $d_u$ allows for a larger margin
  for uncertainties and tracking errors, i.e., the area of $\mathcal{U}_2$ is larger than the area of $\mathcal{U}_1$. Hence
  increasing $d_u$ provides a more robust behavior against uncertainties.
  Notice that, in the case where $c_{\mathcal{B}_i} \notin \mathcal{W}_i$,
  increasing $d_u$ does not guarantee a larger margin. In this case, 
  according to~\eqref{eq:desired_position} and~\eqref{eq:desired_heading}, the
  robot has to be able to adjust its heading holding its position $p_i$. 
  \end{remark}


\section{Simulation results} \label{sec:Simulation results}

We evaluate the proposed \ac{IRBL} method and the integrated system through extensive simulations
in the MRS simulator~\cite{baca2021mrs}, which models the full rigid-body UAV
dynamics.
The methods are evaluated based on the traveled path length $l$, the time to reach the goal $t$, and the average speed over the path $\bar{v}$, which quantify the overall performance of our \ac{IRBL} motion planning method.
We also record and compare the minimum distance to any obstacle over the run $d^o_{\mathrm{min}}$ to demonstrate the safety of our proposed system, especially in dense environments.
At first, we compare the proposed \ac{IRBL} with both RBL~\cite{boldrer2025swarming} and EGO2~\cite{zhou2022swarm} using a single robot
operating in environments with different traversability. Traversability, as
defined in~\cite{de2018challenges}, quantifies the ease of collision-free motion through an environment.
Unlike metrics like mean obstacle density, traversability is calculated using a Monte Carlo approach
by randomly sampling the initial position and direction of motion of
the robot and displacing it in a straight line until it collides with
an obstacle. For each trial $m$, the collision-free travel distance $d_m$ is
recorded. After performing $n$ independent trials, traversability is computed as
\begin{equation}\label{eq:traversability}
\tau = \frac{1}{n}\sum_{m=1}^{n} d_m,
\end{equation}
which corresponds to the average distance traveled by the robot before
encountering an obstacle. Therefore, higher values of $\tau$ indicate an
easy-to-navigate environment, while lower values correspond to more cluttered
spaces.
The success of the single-robot experiment is evaluated on three criteria: violation of the robot's acceleration limits, arrival at the goal, and violation of the robot's radius ($d^o_{\mathrm{min}} < \delta$).
The \ac{SR} for each criterion is then calculated over 10 independent runs. 
These single-robot experiments demonstrate the advantages of using a perception-aware method for 3D navigation in cluttered environments.

After the single robot experiments, we showcase an ablation study to evaluate
the impact of different field-of-view (FoV) configurations and robot dimensions
when operating in a multi-robot setting. We examine three scenarios:  (i) a
circle-crossing scenario without obstacles using \(N=10\) robots, (ii) a
circle-crossing scenario with obstacles using \(N=10\) robots, and (iii) a team
of \(N=5\) robots navigating through a cluttered environment. We use identical
parameters across all the simulations to ensure a fair comparison and
demonstrate the robustness of the proposed \ac{IRBL} method in different conditions.
Finally, we present an analysis evaluating the impact of tracking and measurement errors on collision-free robot motion by varying the parameter $d_u$.
In particular, we show the results obtained from the proposed \ac{IRBL}, by
selecting different values of $d_u$. We included also the case of using a purely
CIRI-based approach to show the advantages of selecting the \ac{CWVD} when surrounded by moving robots.

\subsection{Single-robot comparison}
To rigorously quantify the advantages of our proposed \ac{IRBL} method over established baselines,
we evaluate \ac{IRBL} against RBL~\cite{boldrer2025swarming} and EGO2~\cite{zhou2022swarm} across a suite of diverse environments.
The comparison with RBL is direct and fair; however, additional clarifications are required for EGO2. The simulator used by EGO2 does not model full robot dynamics, which simplifies the robot motion control.
Additionally, the EGO2 method uses inter-robot communication by sharing current position and planned trajectories in multi-robot settings, which
would be unobservable in practice in our communication-free setting.
To ensure methodological fairness and to explicitly capture the inherent limitations of the compared approaches, we restrict our comparison with EGO2 to the single-robot setting.

\begin{figure}[t]
   \includegraphics[width=1.0\columnwidth]{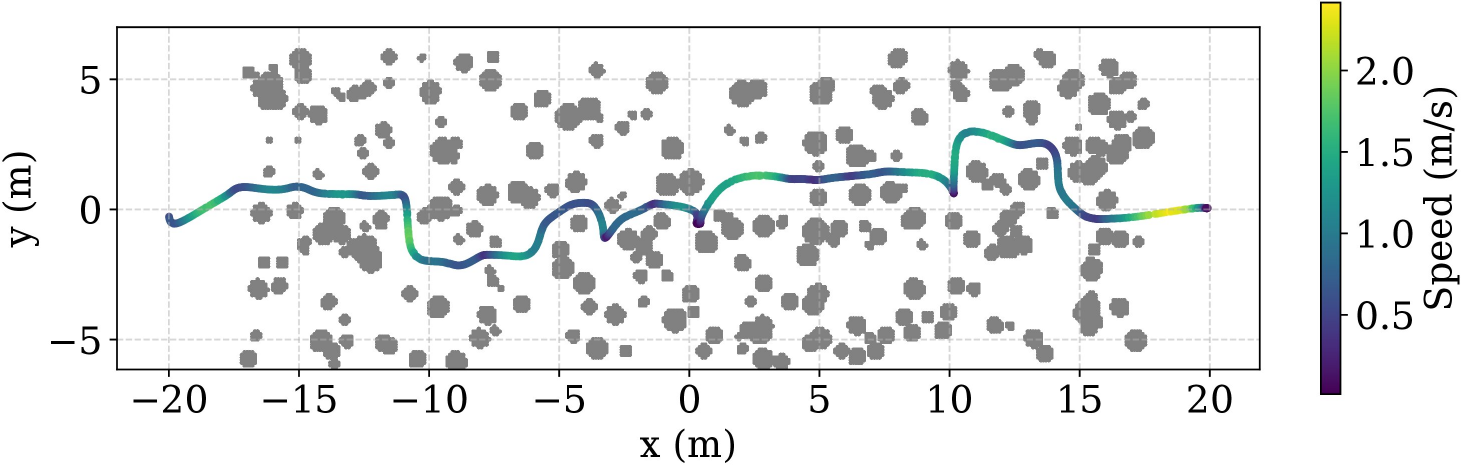} 
   \caption{Top view of the robot's trajectory using \ac{IRBL} in an environment with
   low traversability ($\tau = 1.57$~(m)), where EGO2~\cite{zhou2022swarm} and
   RBL~\cite{boldrer2025swarming} failed to navigate successfully. The color
   scale represents the robot’s velocity, as indicated by the colorbar.}
   \label{fig:exampletrav}
\end{figure}

Table~\ref{tab:single_robot} summarizes the results comparing \ac{IRBL}, \ac{RBL}, and EGO2 on four environments with different traversability $\tau$.
As shown in Table~\ref{tab:single_robot}, \ac{IRBL} achieves a success rate higher than or equal to both RBL and EGO2 across all tested environments. EGO2 often achieves better 
time-to-goal $t$ and average speed $\bar{v}$ but exhibits a
lower success rate, particularly in low-traversability environments. We further observed frequent activation of the emergency-braking mechanism in EGO2, which results
in the violation of the maximum acceleration limit of the robot. We also noticed
frequent deadlocks under dense clutter and occasional safety violations
(collision with obstacles). These findings suggest that optimization-based methods like EGO2 are more failure-prone in low-traversability settings (dense environments) because their planned trajectories need to satisfy multiple, potentially incompatible
feasibility constraints like smoothness, constraints on dynamics, and collision avoidance. In contrast, the iterative re-planning of our proposed system, though simple and focused on obstacle avoidance, practically enhances robustness and overall performance in such challenging environments. 
Figure~\ref{fig:exampletrav} shows the trajectory of a single simulated robot flying using the proposed system and the \ac{IRBL} method in a densely cluttered environment ($\tau=1.57$~(m)). 

\begin{table}[t]
  \centering
  \footnotesize
  \setlength{\tabcolsep}{1pt}
  \setlength{\aboverulesep}{0pt}
  \setlength{\belowrulesep}{0pt}
  \setlength{\cmidrulewidth}{\arrayrulewidth}
  \renewcommand{\arraystretch}{1.0}
  \sisetup{round-mode=places, round-precision=1}
  \begin{tabular}{|l|
                  S[table-format=1.2, round-precision=2]|
                  S[table-format=2.1]|
                  S[table-format=2.1]|
                  S[table-format=1.1]|
                  S[table-format=1.2, round-precision=2]|
                  S[table-format=1.1]|
                  S[table-format=1.1]|
                  S[table-format=1.1]|}
    \hline
    \multirow{2}{*}{}
      & {\multirow{2}{*}{$\tau$ (\unit{\metre})}}
      & {\multirow{2}{*}{$l$ (\unit{\metre})}}
      & {\multirow{2}{*}{$t$ (\unit{\second})}}
      & {\multirow{2}{*}{$\bar{v}$ (\unit{\metre\per\second})}}
      & {\multirow{2}{*}{$d^o_{\min}$ (\unit{\metre})}}
      & \multicolumn{3}{c|}{\ac{SR}} \\ \cmidrule{7-9}
      & & & & & & {a} & {g} & {d} \\
    \hline\hline
    \multirow{4}{*}{EGO2~\cite{zhou2022swarm}}
      & 3.566 & 31.37 & 15.17 & 2.07 & 0.486 & 0.4 & 1   & 1   \\
      & 2.036 & 36.01 & 20.32 & 1.82 & 0.460 & 0   & 1   & 1   \\ 
      & 1.142 & 35.25 & 23.20 & 1.51 & 0.520 & 0.2 & 0.2 & 0.8 \\
      & 0.984 & {--}  & {--}  & {--} & {--}  & 0   & 0   & 0.8 \\ \hline\hline
    \multirow{4}{*}{RBL~\cite{boldrer2025swarming}}
      & 3.566 & 36.31 & 52.90 & 0.68 & 0.450 & 1   & 0.6 & 1   \\
      & 2.036 & 34.56 & 71.92 & 0.48 & 0.430 & 1   & 0.4 & 1   \\
      & 1.142 & {--}  & {--}  & {--} & {--}  & 1   & 0   & 1   \\
      & 0.984 & {--}  & {--}  & {--} & {--}  & 1   & 0   & 1   \\ \hline\hline
    \multirow{4}{*}{\shortstack{\ac{IRBL}\\(full FoV)}}
      & 3.566 & 33.08 & 25.36 & 1.33 & 0.460 & 1   & 1   & 1   \\
      & 2.036 & 36.54 & 41.85 & 0.90 & 0.390 & 1   & 1   & 1   \\
      & 1.142 & 39.03 & 57.04 & 0.70 & 0.373 & 1   & 1   & 1   \\
      & 0.984 & 42.68 & 53.78 & 0.81 & 0.404 & 1   & 1   & 1   \\ \hline
  \end{tabular}
  \vspace{5pt}
  \caption{Comparison between proposed \ac{IRBL} and state-of-the-art \ac{RBL}~\cite{boldrer2025swarming}, and EGO2~\cite{zhou2022swarm} methods.
    Simulation with $N=1$ robot and $\delta_{i}=0.2$~m in $4$ different cluttered environments quantified by the
    traversability $\tau$.
    The methods are evaluated on the traveled path length $l$, time to reach the goal $t$, average speed over the path $\bar{v}$, and the minimum distance to any obstacle $d^o_{\mathrm{min}}$.
   The \acf{SR} is calculated over $10$ repeated runs where columns a, g, and d denote the criteria for violation of acceleration limits, arrival at goal, and violation of robot radius, respectively.
  }
  \label{tab:single_robot}
\end{table}


\subsection{Multi-robot comparison} 
\begin{figure*}[t]
  \centering
  \begin{subfigure}[b]{0.32\textwidth}
    \centering
    \includegraphics[width=\linewidth]{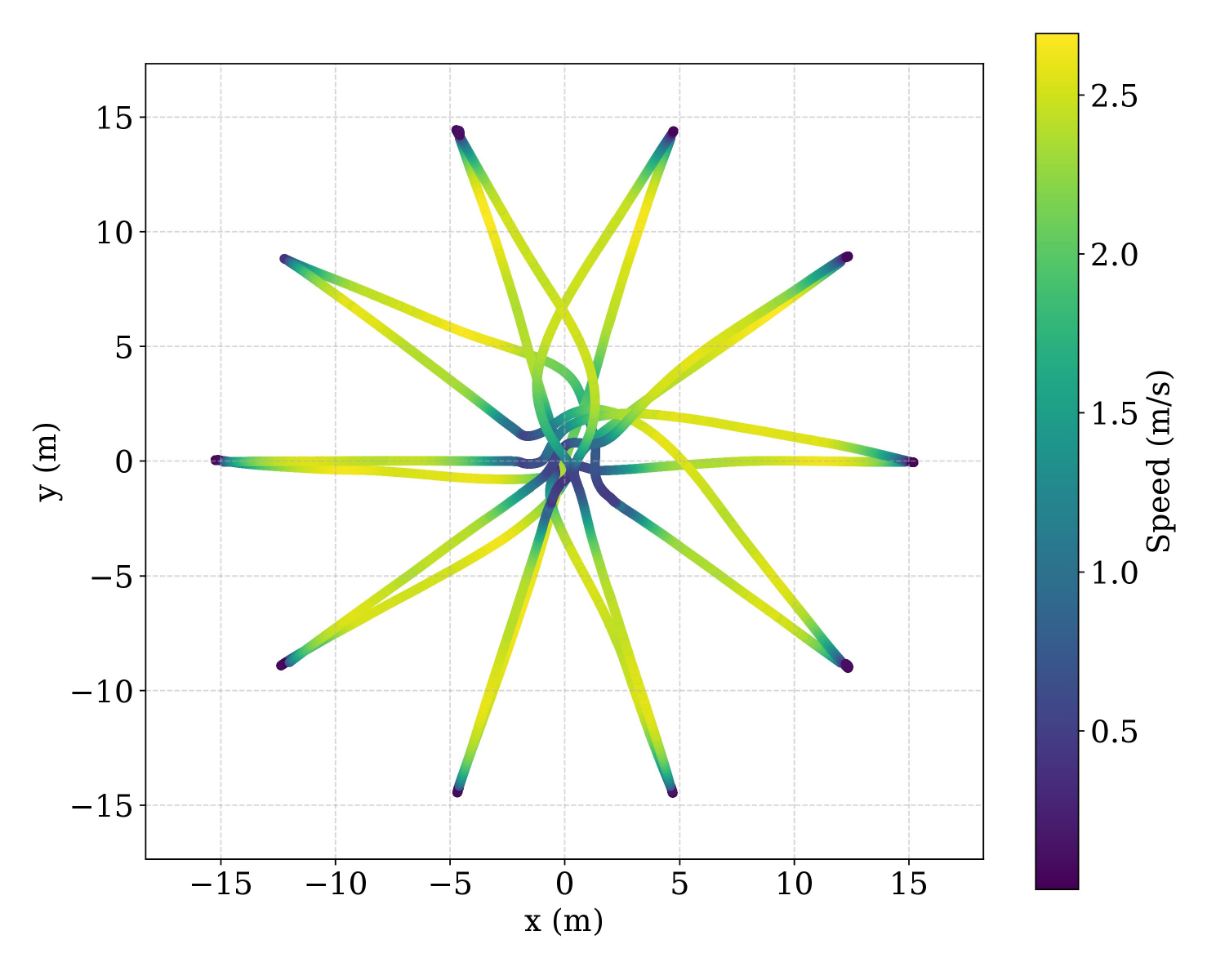}
    \caption{Cross, no obstacles ($N=10$)}
    \label{fig:sim_circle}
  \end{subfigure}\hfill
  \begin{subfigure}[b]{0.32\textwidth}
    \centering
    \includegraphics[width=\linewidth]{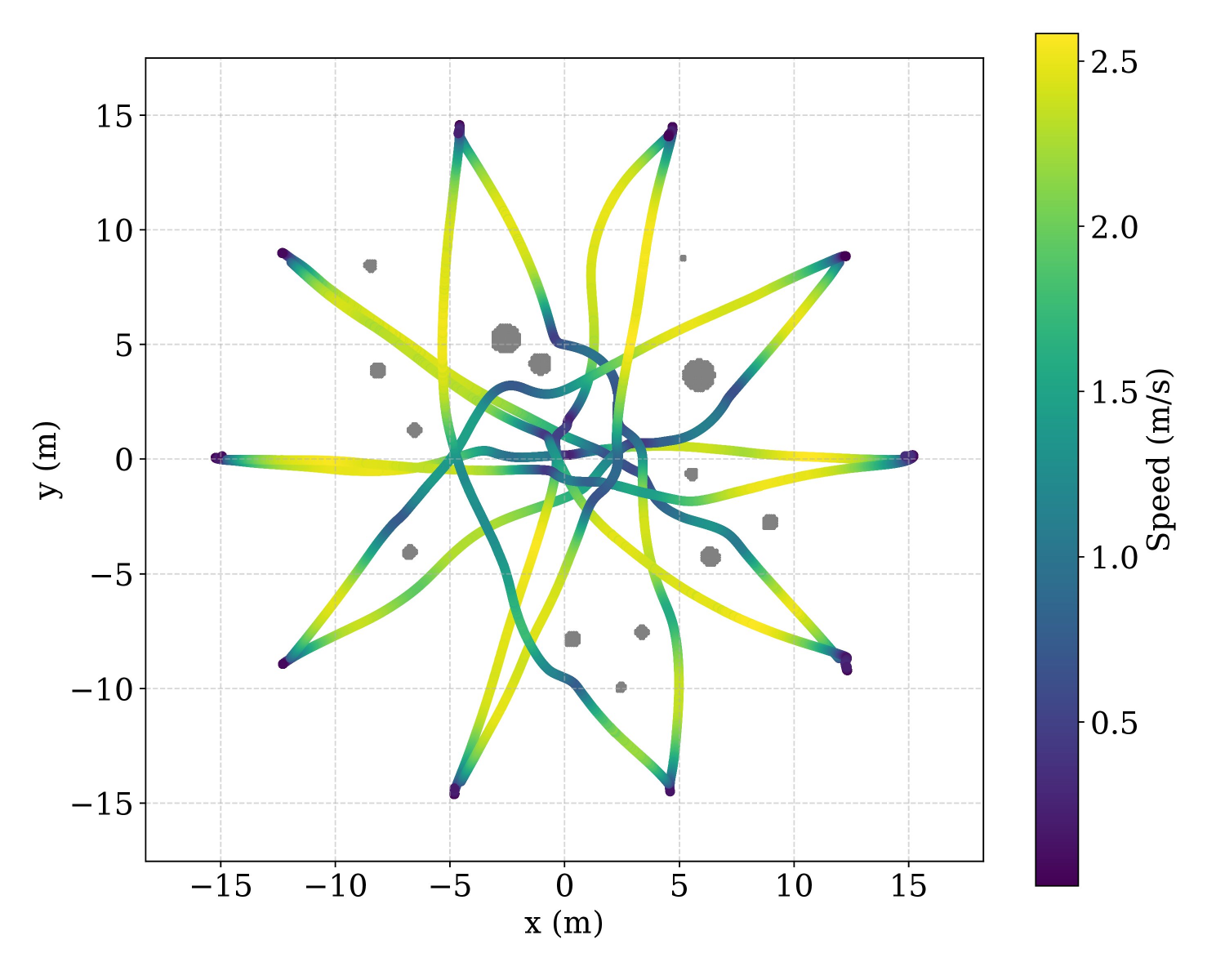}
    \caption{Cross, with obstacles ($N=10$)}
    \label{fig:sim_obstacles}
  \end{subfigure}\hfill
  \begin{subfigure}[b]{0.32\textwidth}
    \centering
    \includegraphics[width=\linewidth]{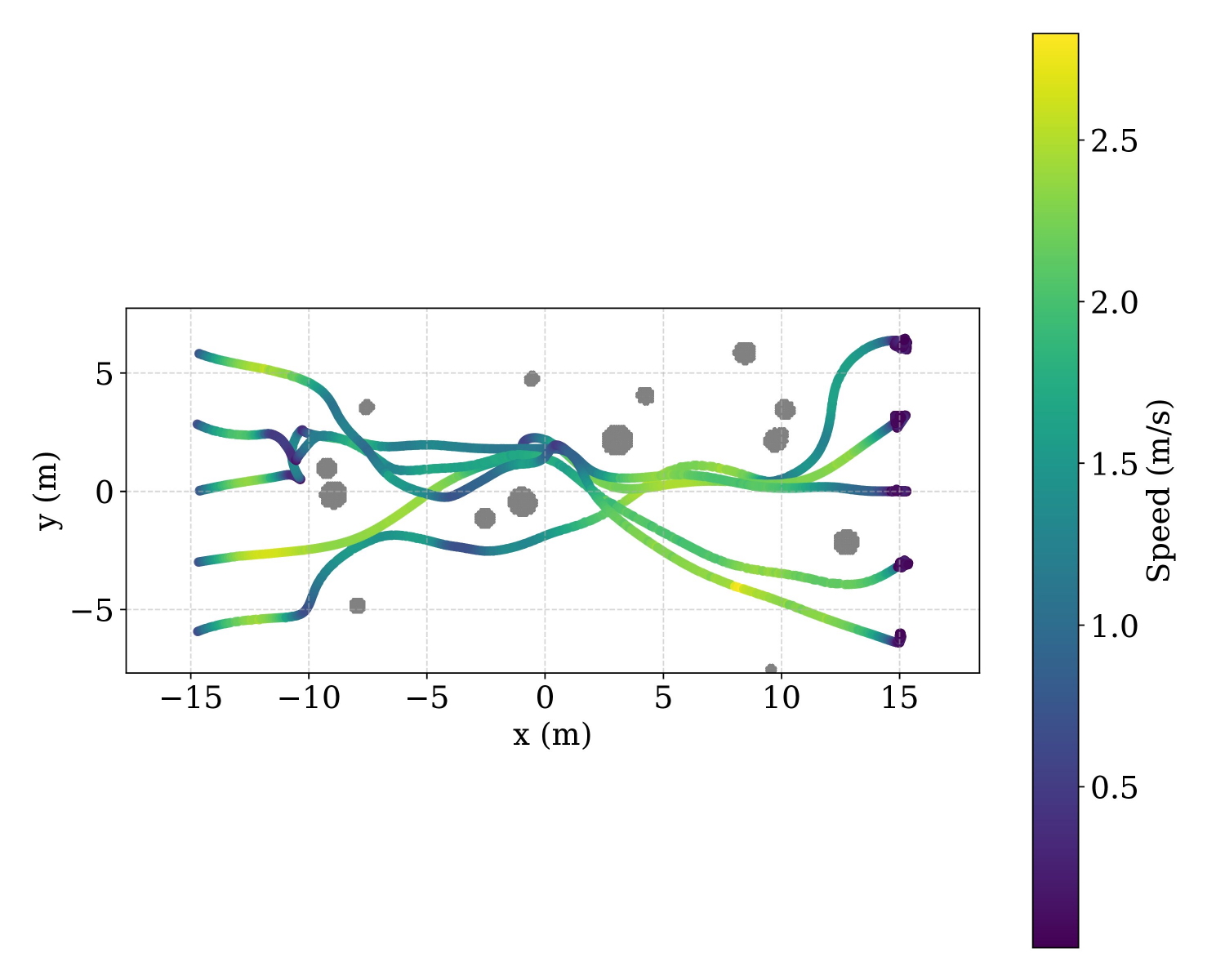}
    \caption{Forest-like environment ($N=5$)}
    \label{fig:sim_forest}
  \end{subfigure}
  \caption{Top view of the robots' trajectories for \ac{IRBL} with half FoV
    $\{180, 180, -90\}$~(deg) and $\delta_i = 0.2$~(m). The color indicates the
    velocities according to the colorbars.}
  \label{fig:sim}
\end{figure*}

This section evaluates the performance of our proposed decentralized 3D navigation system in three multi-robot scenarios under different field-of-view (FoV) configurations and compares the performance with RBL~\cite{boldrer2025swarming}.
Notice that, as reported in Table~\ref{tab:sota}, the only open source methods that do not rely on inter-robot communication are RBL~\cite{boldrer2025swarming} and
PACNAV~\cite{ahmad2022pacnav}. Since RBL outperforms PACNAV (as demonstrated in~\cite{boldrer2025swarming}), we picked the former as the baseline.
Unless otherwise stated (see the sensitivity analysis on $d_u$ in Section~\ref{sec:sensitivity}), for all simulations we select the following parameters: $d_u=0.3$~(m), $\beta^D=0.5$, $\epsilon=0.5$,
$d_1=d_2=d_3=d_4=1.0$~(m), $r_{s,i}=5.0$~(m), and $k_{\mathrm{wp}}=k_{\beta}=1$. We consider different fields of view
FoV = \{180,59,-20\}, \{180,180,-90\}, \{180,360,0\}~(deg),
and different robot sizes $\delta_i=[0.2,\,0.5,\,1.0]$~(m). Figure~\ref{fig:sim} shows the trajectories obtained for each scenario with $\delta_i=0.2$~(m) and $\mathrm{FoV}=\{180,180,-90\}~(\mathrm{deg})$. For each scenario, the quantitative results are reported in Tables~\ref{tab:sim1}, \ref{tab:sim2}, and \ref{tab:sim3}. The results consistently demonstrate a clear improvement of \ac{IRBL} over RBL, with higher average speed $\bar{v}$ and shorter travel time $t$ across all scenarios and different \ac{FOV} settings.
The success of our proposed \ac{IRBL} method in navigating through densely cluttered environments with multiple robots also underscores the safety of the proposed decentralized system. A single set of parameters was used across all experiments despite substantial variations in robot size $\delta$, sensing configuration (\ac{FOV}), and environmental complexity ($\tau$). While performance naturally degrades as
traversability decreases (e.g., with robot size increases) and tends to improve with larger fields-of-view, the method maintains stable and consistent behavior across all tested conditions. These observations indicate that the proposed approach is not overly sensitive to parameter selection and generalizes well across different robot morphologies, sensing capabilities, and levels of environmental complexity.

\begin{table}[t]
  \centering

  \footnotesize
  \renewcommand{\tabcolsep}{0.04cm} 
\begin{tabular}{|c|c|c|c|c|c|c|} \hline
 & $\tau$ (m) & $\delta_i$ (m) & $l$ (m) & $t$ (s) & $\bar{v}$ (m/s) & $d_{\min}$ (m) \\ \hline\hline

\multirow{3}{*}{\ac{IRBL} (lim.FoV)}
& - & 0.2 & 33.14 & 23.43 & 1.41 & 0.737 \\ 
& - & 0.5 & 33.99 & 26.72 & 1.26 & 1.155 \\
& - & 1.0 & 35.51 & 33.95 & 1.04 & 2.17 \\ \hline\hline

\multirow{3}{*}{\ac{IRBL} (half FoV)}
& - & 0.2 & 33.56 & 22.14 & 1.50 & 0.729 \\
& - & 0.5 & 34.30 & 25.09 & 1.35 & 1.402 \\
& - & 1.0 & 35.14 & 27.62 & 1.26 & 2.092 \\ \hline\hline

\multirow{3}{*}{\ac{IRBL} (full FoV)}
& - & 0.2 & 34.30 & 20.48 & 1.67 & 0.709 \\
& - & 0.5 & 34.35 & 24.95 & 1.36 & 1.176 \\
& - & 1.0 & 35.62 & 28.67 & 1.23 & 2.218 \\ \hline\hline

\multirow{3}{*}{RBL (2d)}
& - & 0.2 & 34.17 & 41.58 & 0.83 & 0.827 \\
& - & 0.5 & 36.67 & 71.66 & 0.51 & 1.385 \\
& - & 1.0 & 38.39 & 65.58 & 0.59 & 2.373 \\ \hline

\end{tabular}

  \vspace{10pt}
  \caption{Simulation results with $N=10$ robots of radius $\delta_{i} = [0.2, 0.5, 1.0]$ in a crossing circle scenario
  without obstacles.
  The methods are evaluated on the traveled path length $l$, time to reach the goal $t$, average speed over the path $\bar{v}$, and the minimum distance to any obstacle or robot $d_{\min}$.
  The \ac{IRBL} is evaluated on limited FoV \{180,59,-20\}~(deg), half FoV \{180,180,-90\}~(deg), and full FoV \{180,360,0\}~(deg) and compared with the two-dimensional case
\{360,0,0\}~(deg) of \ac{RBL}~\cite{boldrer2025swarming}.}
  \label{tab:sim1}
\end{table}

\begin{table}[t]
\centering
\renewcommand{\tabcolsep}{0.04cm} 
\footnotesize
\begin{tabular}{|c|c|c|c|c|c|c|} \hline
 & $\tau$ (m) & $\delta_i$ (m) & $l$ (m) & $t$ (s) & $\bar{v}$ (m/s) & $d_{\min}$ (m) \\ \hline\hline

\multirow{3}{*}{\ac{IRBL} (lim.\ FoV)}
& 6.56 & 0.2 & 32.46 & 23.24 & 1.40 & 0.661 \\
& 4.60 & 0.5 & 35.83 & 33.07 & 1.08 & 1.165 \\
& 2.21 & 1.0 & 38.54 & 55.06 & 0.69 & 2.199 \\ \hline\hline

\multirow{3}{*}{\ac{IRBL} (half FoV)}
& 6.56 & 0.2 & 32.70 & 23.74 & 1.37 & 0.878 \\ 
& 4.60 & 0.5 & 35.99 & 34.47 & 1.04 & 1.178 \\
& 2.21 & 1.0 & 38.17 & 43.75 & 0.86 & 2.119 \\ \hline\hline

\multirow{3}{*}{\ac{IRBL} (full FoV)}
& 6.56 & 0.2 & 32.97 & 23.91 & 1.37 & 0.791 \\
& 4.60 & 0.5 & 35.38 & 31.03 & 1.14 & 1.224 \\
& 2.21 & 1.0 & 37.31 & 38.36 & 0.96 & 2.074 \\ \hline\hline

\multirow{3}{*}{RBL (2d)}
& 6.56 & 0.2 & 33.60 & 41.87 & 0.80 & 0.924 \\
& 4.60 & 0.5 & 38.01 & 58.86 & 0.64 & 1.329 \\
& 2.21 & 1.0 & - & - & - & - \\ \hline
\end{tabular}




  \vspace{10pt}
  \caption{Simulation results with $N=10$ robots of radius $\delta_{i} = [0.2, 0.5, 1.0]$ in a crossing circle scenario
  with obstacles.
  The methods are evaluated on the traveled path length $l$, time to reach the goal $t$, average speed over the path $\bar{v}$, and the minimum distance to any obstacle or robot $d_{\min}$.
  The \ac{IRBL} is evaluated on limited FoV \{180,59,-20\}~(deg), half FoV \{180,180,-90\}~(deg), and full FoV \{180,360,0\}~(deg) and compared with the two-dimensional case
\{360,0,0\}~(deg) of \ac{RBL}~\cite{boldrer2025swarming}.}
  \label{tab:sim2}
\end{table}

\begin{table}[t]
\centering
\footnotesize
\renewcommand{\tabcolsep}{0.04cm} 
\begin{tabular}{|c|c|c|c|c|c|c|} \hline
 & $\tau$ (m) & $\delta_i$ (m) & $l$ (m) & $t$ (s) & $\bar{v}$ (m/s) & $d_{\min}$ (m) \\ \hline\hline

\multirow{3}{*}{\ac{IRBL} (lim.\ FoV)}
& 8.67 & 0.2 & 34.58 & 26.83 & 1.28 & 0.976 \\
& 7.01 & 0.5 & 36.51 & 30.94 & 1.18 & 1.723 \\
& 4.17 & 1.0 & 39.60 & 52.31 & 0.76 & 2.216 \\ \hline\hline

\multirow{3}{*}{\ac{IRBL} (half FoV)}
& 8.67 & 0.2 & 35.10 & 26.52 & 1.33 & 1.151 \\ 
& 7.01 & 0.5 & 36.94 & 30.25 & 1.22 & 1.768 \\
& 4.17 & 1.0 & 37.31 & 34.77 & 1.07 & 2.235 \\ \hline\hline

\multirow{3}{*}{\ac{IRBL} (full FoV)}
& 8.67 & 0.2 & 35.49 & 26.48 & 1.33 & 1.192 \\
& 7.01 & 0.5 & 36.57 & 25.66 & 1.42 & 1.492 \\
& 4.17 & 1.0 & 38.90 & 34.09 & 1.14 & 2.278 \\ \hline\hline

\multirow{3}{*}{RBL (2d)}
& 8.67 & 0.2 & 34.75 & 38.03 & 0.91 & 1.372 \\
& 7.01 & 0.5 & 39.29 & 37.16 & 1.06 & 1.566 \\
& 4.17 & 1.0 & 39.36 & 74.05 & 0.53 & 2.321 \\ \hline

\end{tabular}
  \vspace{10pt}
  \caption{Simulation results with $N=5$ robots of radius $\delta_{i} = [0.2, 0.5, 1.0]$ in a forest-like environment with varying traversability $\tau$.
  The methods are evaluated on the traveled path length $l$, time to reach the goal $t$, average speed over the path $\bar{v}$, and the minimum distance to any obstacle or robot $d_{\min}$.
  The \ac{IRBL} is evaluated on limited FoV \{180,59,-20\}~(deg), half FoV \{180,180,-90\}~(deg), and full FoV \{180,360,0\}~(deg) and compared with the two-dimensional case
\{360,0,0\}~(deg) of \ac{RBL}~\cite{boldrer2025swarming}.}
  \label{tab:sim3}
\end{table}








\subsection{Multi-robot sensitivity analysis}\label{sec:sensitivity}
Finally, we present the results of 40 simulations performed with different
values of the parameter $d_u \in \{0.0,\,0.5,\,1.0\}$~(m), together with the
case of a pure CIRI-based planner, in which the waypoint to follow is
re-projected inside the safe set. We focus on the most critical scenario in
terms of safety, i.e., the circle-crossing scenario without obstacles. The
results are in line with the considerations made in Remark~\ref{re:safety}. In
particular, selecting $d_u = 0.0$~(m) yields a success rate of $SR = 0.98$,
meaning that safety is not always achieved. Conversely, selecting $d_u \in
\{0.5,\,1.0\}$~(m) is sufficient to obtain a safe behavior in all runs, with a
minimum inter-robot distance of $d_{\min} \in \{0.89,\,1.23\}$~(m),
respectively. Finally, the pure CIRI-based planner results in a more greedy
behavior of the robots, but at the price of a lower success rate, which is
unacceptable for safety-critical applications. The complete results are reported
in Table~\ref{tab:sens}.

\begin{table}[t]
  \centering
  \footnotesize
  \renewcommand{\tabcolsep}{0.07cm} 
\begin{tabular}{|c|c|c|c|c|c|c|} \hline
  &  $\delta_i$ (m) & $l$ (m) & $t$ (s) & $\bar{v}$ (m/s) & $d_{\min}$ (m) & SR \\ \hline\hline

  \multirow{1}{*}{CIRI-based}
  & 0.3 & 37.78 & 37.79  &1.075  & 0.016 & 0.11  \\ 
  \hline
  \multirow{1}{*}{\ac{IRBL} ($d_u=0$)}
  & 0.3 & 37.81 & 60.45 & 0.75 & 0.597 & 0.98 \\ 
  \hline
\multirow{1}{*}{\ac{IRBL} ($d_u=0.5$)}
  & 0.3 & 37.85 & 45.72 & 0.82 & 0.89 & 1.0 \\ 
  \hline
  \multirow{1}{*}{\ac{IRBL} ($d_u=1.0$)}
    & 0.3 & 38.39 & 46.52 & 0.81 & 1.23 & 1.0 \\ 
    \hline

\end{tabular}
  \vspace{10pt}
  \caption{Simulation results with $N=10$ robots in a crossing circle scenario
  without obstacles. We report the averages among 10 simulations of the length of
  the path $l$, of the time to reach the goal $t$, of the average speed $\bar{v}$, 
  the minimum distance between the robots $d_{\min}$, and the success rate (SR),
  which is computed for each UAV across all simulations. We considered the limited
  FoV \{180,59,-20\}~(deg), with $d_u=[0.0,\,0.5,\,1.0]$~(m) and a pure
  CIRI-based with a pure waypoint re-projection inside the safe set.}
  \label{tab:sens}
\end{table}

\section{Experimental Results} \label{sec:Experimental results}
We deploy and evaluate the proposed perception-aware \ac{IRBL} method and the decentralized multi-robot system on real aerial robotic platforms.
Each platform is equipped with a Livox MID-360 LiDAR with a field-of-view specified as $\{180, 59, -20\}$ (deg).
The Livox sensor is central to our pipeline: (i) it supplies measurements for real-time odometry estimation
via Point-LIO~\cite{he2023point}; (ii) it enables the construction of a 3D
environmental map, allowing for obstacles
avoidance and replanning; and (iii) it facilitates the relative localization of neighboring UAVs by detecting
reflective markers mounted on their frames.
The relative positions of the neighbors are obtained by fusing observations using a Kalman filter~\cite{10802770,10758263}.

\begin{figure*}[t]
 \centering

 \setlength{\tabcolsep}{0.05em}
 \begin{tabular}{ccc}
  \includegraphics[width=.69\columnwidth]{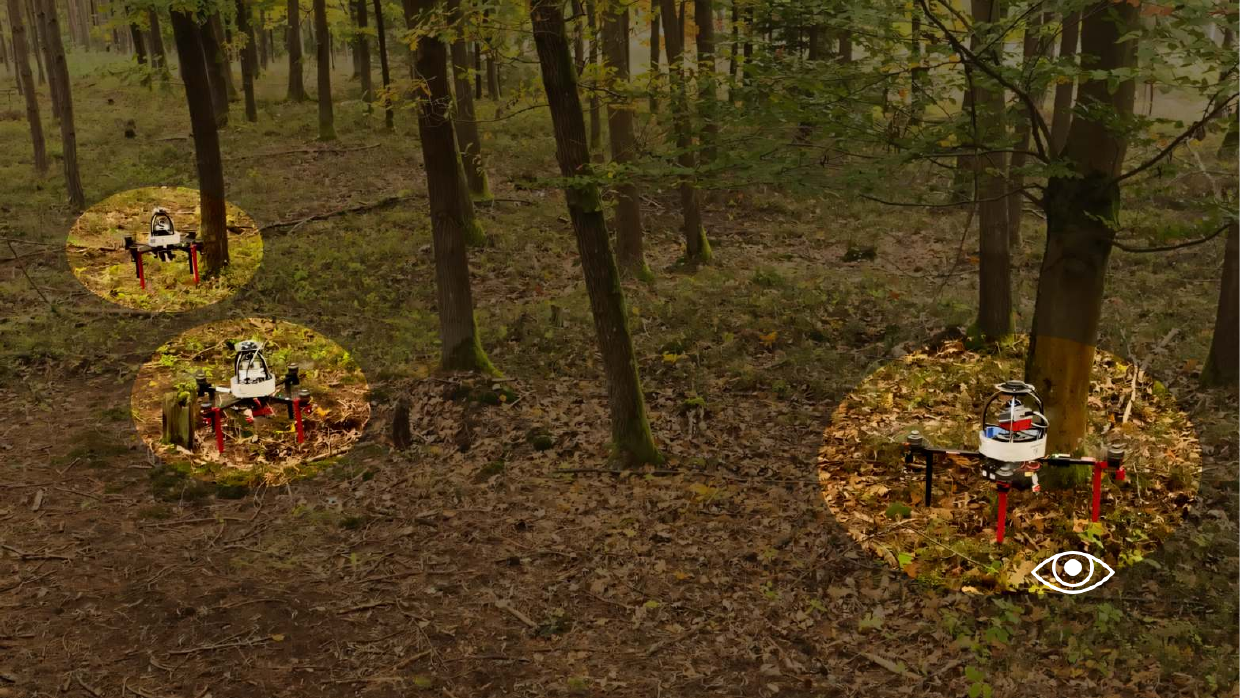} & 
  \includegraphics[width=.69\columnwidth]{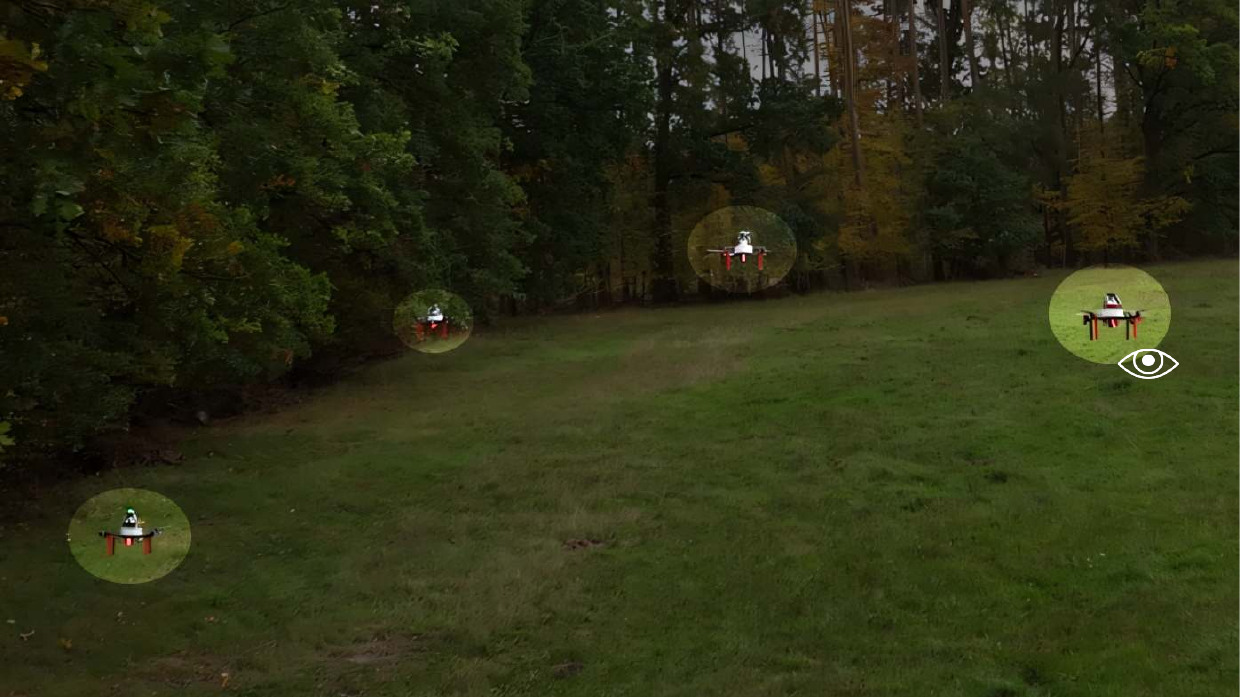} & 
 \includegraphics[width=.69\columnwidth]{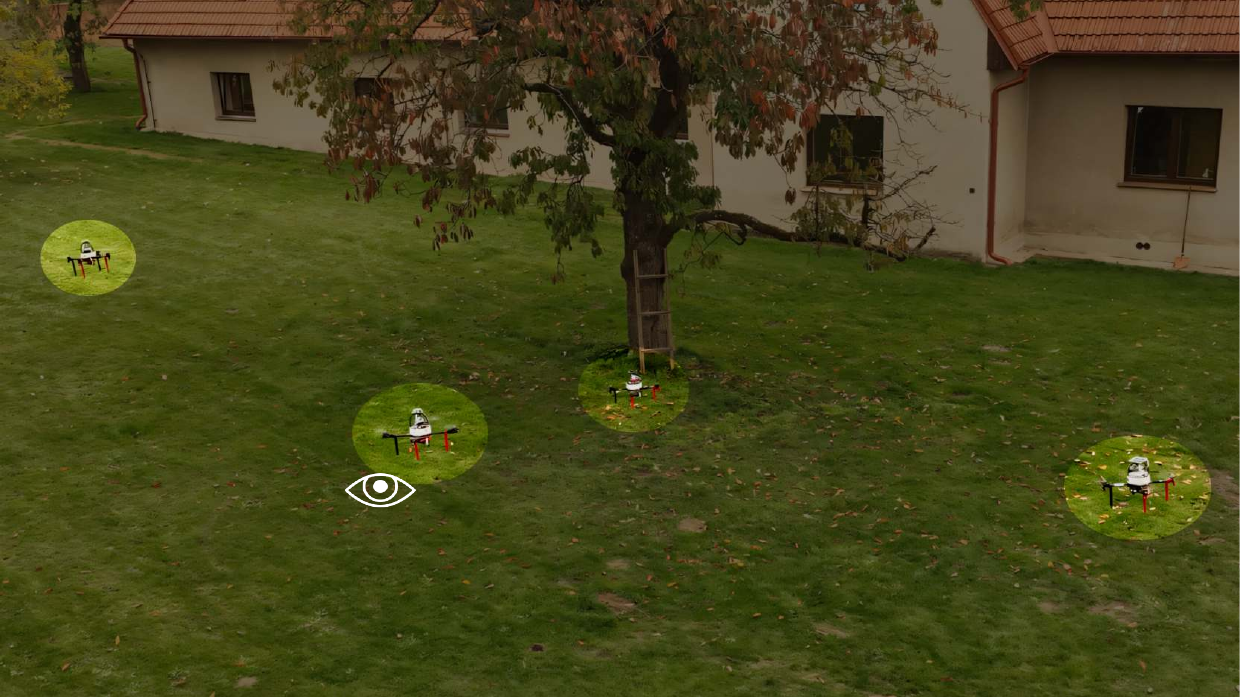}\\ 
   \includegraphics[width=.69\columnwidth]{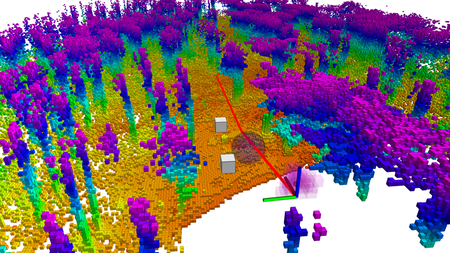} & 
   \includegraphics[width=.69\columnwidth]{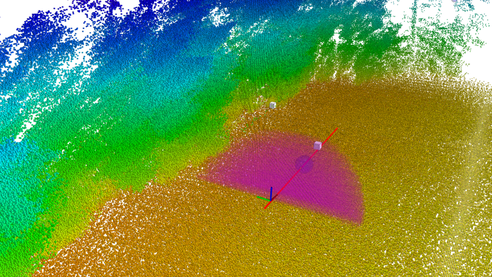} & 
  \includegraphics[width=.69\columnwidth]{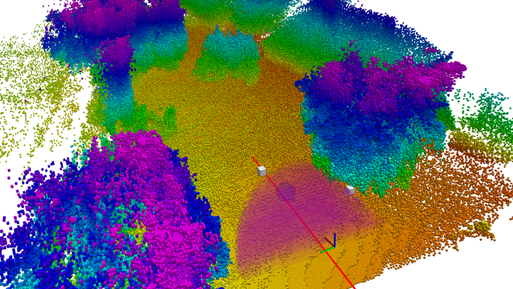}\\ 
   (a) & (b) & (c)
   \end{tabular}
   \caption{
  Different experimental scenarios. The top row contains snapshots from the
  camera, while the bottom row shows the corresponding snapshots extracted from
  RViz for the same experiment. Similarly to Figure~\ref{fig:rviz}, the robot
  pose is represented by an axis reference frame. The set $\mathcal{W}_i$ is
  depicted as magenta dots, and the centroid $c_{\mathcal{B}_i}$ as an orange
  sphere. The planned path is shown as a red line. The active waypoint $wp_i$
  lies on the planned path and is represented by a purple sphere. The grey boxes
  represent the positions of the other robots. The colored point cloud
  represents obstacles, where different colors correspond to different
  $z$-values. The ``eye'' symbol indicates the robot whose viewpoint is
  currently shown in the bottom RViz visualization.
   } 
   \label{fig:exp}
\end{figure*}
\begin{figure*}[t]
 \centering
 \setlength{\tabcolsep}{0.05em}
 \begin{tabular}{ccccc}
   \includegraphics[width=.63\columnwidth]{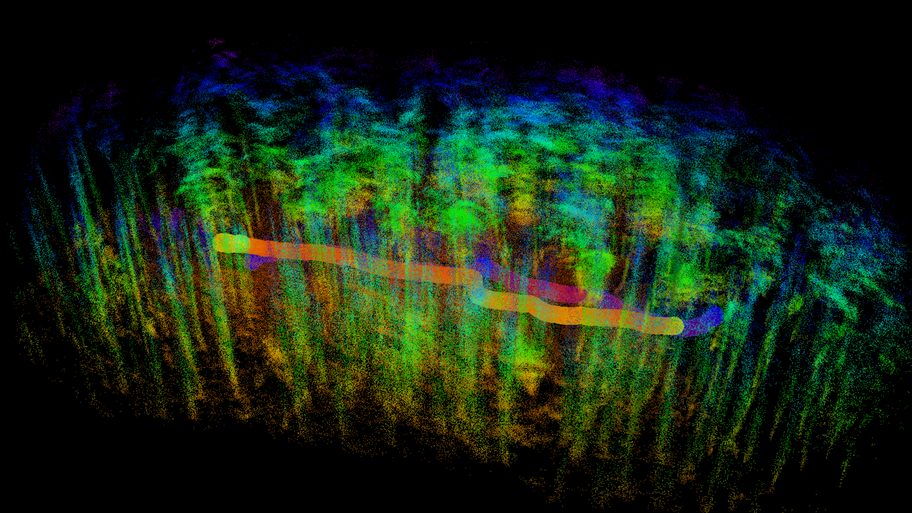} &
   \includegraphics[width=.63\columnwidth]{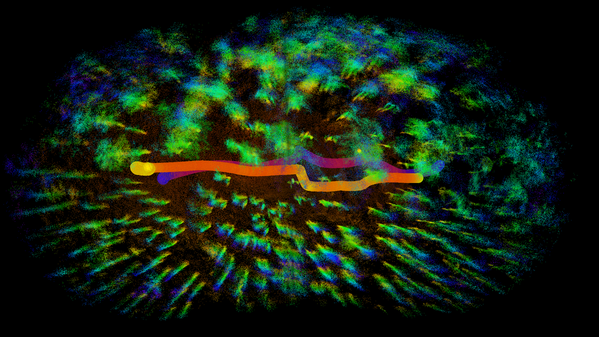} &
   \includegraphics[width=.63\columnwidth]{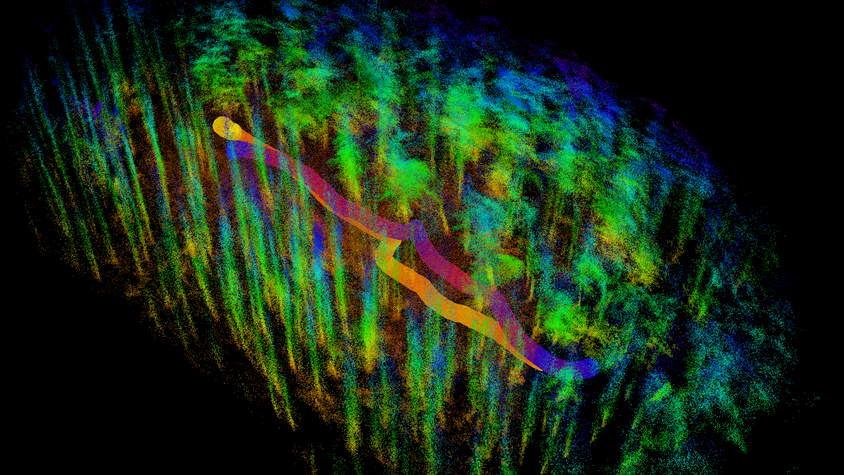} &
   \includegraphics[width=.09\columnwidth]{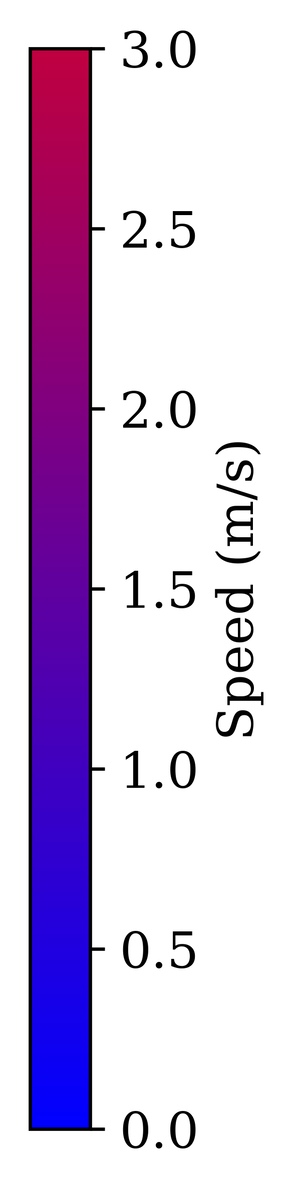} &
\includegraphics[width=.09\columnwidth]{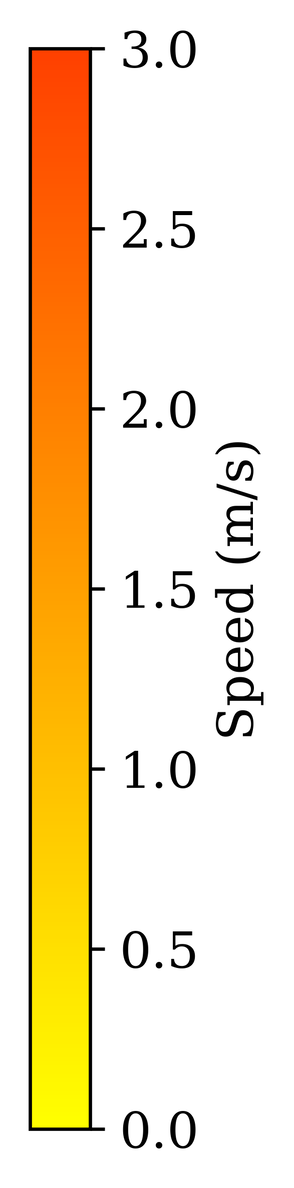}
   \end{tabular}
   \caption{
     Different viewpoints from experiment scenario \#1 of Table~\ref{tab:exp1}. We depict
     the trajectories followed by the two robots during the mission. The robots'
     velocities can be inferred from the color-bars on the right-hand side.
     Different colors in the point cloud indicate different $z$-values.}
  \label{fig:exp_forest}
\end{figure*}

The proposed perception-aware navigation system (see Figure~\ref{fig:block}) runs on an
Intel NUC (i7-10750H CPU and 16~GB RAM) mounted on each robot. We successfully
evaluated the proposed approach across multiple scenarios. Figure~\ref{fig:exp}
showcases representative environments: (a) a dense forest, (b) an open area
adjacent to a forest, and (c) a backyard near a residential structure with
trees. Across all experiments, we retain identical parameter settings,
demonstrating robustness to environmental variation and to different
team sizes.
In particular, we select $r_{s,i}=5.0$~(m), $d_1=d_2=d_3=d_4=1.0$~(m),
$\beta^D=0.5$, $d_u=0.5$~(m),
$\mathrm{FoV}=\{180, 59, -20\}~(\mathrm{deg})$, and $\delta_i=0.7$~(m).
Quantitative performance metrics for the scenarios illustrated in
Figure~\ref{fig:exp} are summarized in Table~\ref{tab:exp1}.
Figure~\ref{fig:exp_forest} shows the trajectories
followed by two robots in the scenario \#1 as reported in Table~\ref{tab:exp1}. 
More experimental results are provided in the multimedia material.

\begin{table}[t]
\footnotesize
\centering
\renewcommand{\tabcolsep}{0.12cm} 
\begin{tabular}{|c|c|c|c|c|c|c|c|} \hline
scenario & uav & $\tau$ (m) & $l$ (m) & $t$ (s) & $\bar{v}$ (m/s) & $v_{\max}$ (m/s) \\ \hline\hline

\multirow{2}{*}{\#1 (40m)}
& 1 & 3.10 & 45.77 & 23.00 & 1.99 & 2.60 \\ 
& 2 & 3.10 & 41.98 & 23.45 & 1.79 & 2.85 \\ \hline\hline

\multirow{3}{*}{\#2 (80m)}
& 1 & 16.02 & 86.97 & 35.07 & 2.48 & 2.79 \\
& 2 & 16.02 & 91.02 & 45.51 & 2.00 & 2.69 \\
& 3 & 16.02 & 94.34 & 53.00 & 1.78 & 2.81 \\ \hline\hline

\multirow{3}{*}{\#3 (40m)}
& 1 & 6.73 & 46.06 & 23.03 & 2.00 & 3.32 \\ 
& 2 & 6.73 & 41.83 & 29.25 & 1.43 & 2.90 \\
& 3 & 6.73 & 46.21 & 32.54 & 1.42 & 2.79 \\ \hline
\end{tabular}
  \vspace{10pt}
  \caption{
    Quantitative results for multi-robot experiments in the scenarios illustrated in Figure~\ref{fig:exp}-(a), (b) and
  (c), respectively. All the scenarios involve aerial robots starting close to each other and going to individual goals that are also in close vicinity to each other.
  The system is evaluated based on the traveled path length $l$, the time to reach the goal $t$, and the average and maximum speeds along the path ($\bar{v}, v_{\mathrm{max}}$).
}
  \label{tab:exp1}
\end{table}

\section{Conclusions, limitations and future work} \label{sec:Conclusions} 
We presented a novel multi‑robot implicit coordination framework for GNSS‑denied
environments that operates solely with onboard anisotropic 3D‑LiDAR sensing. Our
approach does not rely on inter-robot communication and it explicitly handles
limited and directional sensor field-of-view. Simulation studies indicate that
the proposed approach outperforms state‑of‑the‑art baselines, particularly in
low‑traversability scenarios. Moreover, we show strong generalization across
robots of different sizes, sensing profiles, and environmental conditions.
Finally, real‑world deployments validate the effectiveness of the framework,
demonstrating robustness to parameters' selection across different scenarios.

The major limitation of the proposed approach lies in the selection of the
parameter $d_u$, which accounts for tracking errors and measurement
uncertainties. In practice, the uncertainty bound must be tuned with care to
prevent collisions. Its admissible range is not solely dictated by measurement
uncertainty, but it is also a function of the controller design, and the underlying
dynamics. Choosing a value of $d_u$ that is too large can erode performance,
whereas selecting it too small can jeopardize system safety. Hence an initial
tuning of $d_u$ is required to find a balance between performance and safety.


In our future research, we will extend our estimation pipeline beyond relative
positions to include additional observable states of nearby robots, and leverage
these estimates to further improve implicit coordination. We will validate the
proposed algorithm on a larger number of smaller, more agile platforms to enable
operation in denser forests and other challenging environments. Lastly, we will
explore adding minimal explicit communication to support high-level mission
coordination while preserving the decentralized nature of the proposed method. 




\bibliography{reference}

\end{document}